%% file: main.tex
\documentclass{article}

\usepackage[utf8]{inputenc}
\usepackage{microtype}
\usepackage{graphicx}
\usepackage{subcaption}
\usepackage{booktabs} 

\usepackage{hyperref}
\usepackage{fancyhdr}
\usepackage{tocloft}
\usepackage{etoc}

\usepackage{algorithm}
\usepackage{algorithmic}

\usepackage{float}

\usepackage[preprint]{icml2026}

\usepackage{amsmath}
\usepackage{amssymb}
\usepackage{mathtools}
\usepackage{amsthm}

\usepackage{multirow}

 \usepackage{xcolor,colortbl}
\usepackage{tcolorbox}
\tcbuselibrary{breakable,skins}

\usepackage{gradient-text}
\usepackage{xspace}

\usepackage{booktabs}
\usepackage{makecell}

\usepackage{pifont}

\usepackage{enumitem}

\usepackage[capitalize,noabbrev]{cleveref}

\theoremstyle{plain}

\theoremstyle{definition}

\theoremstyle{remark}

\usepackage[textsize=tiny]{todonotes}

\newcommand{\ours}{\gradientRGB{AblateCell}{53,93,203}{10,10,80}\xspace}

\newcommand{\titleLogo}{%
    \raisebox{-3.1mm}
    {\includegraphics[width=0.06\textwidth]{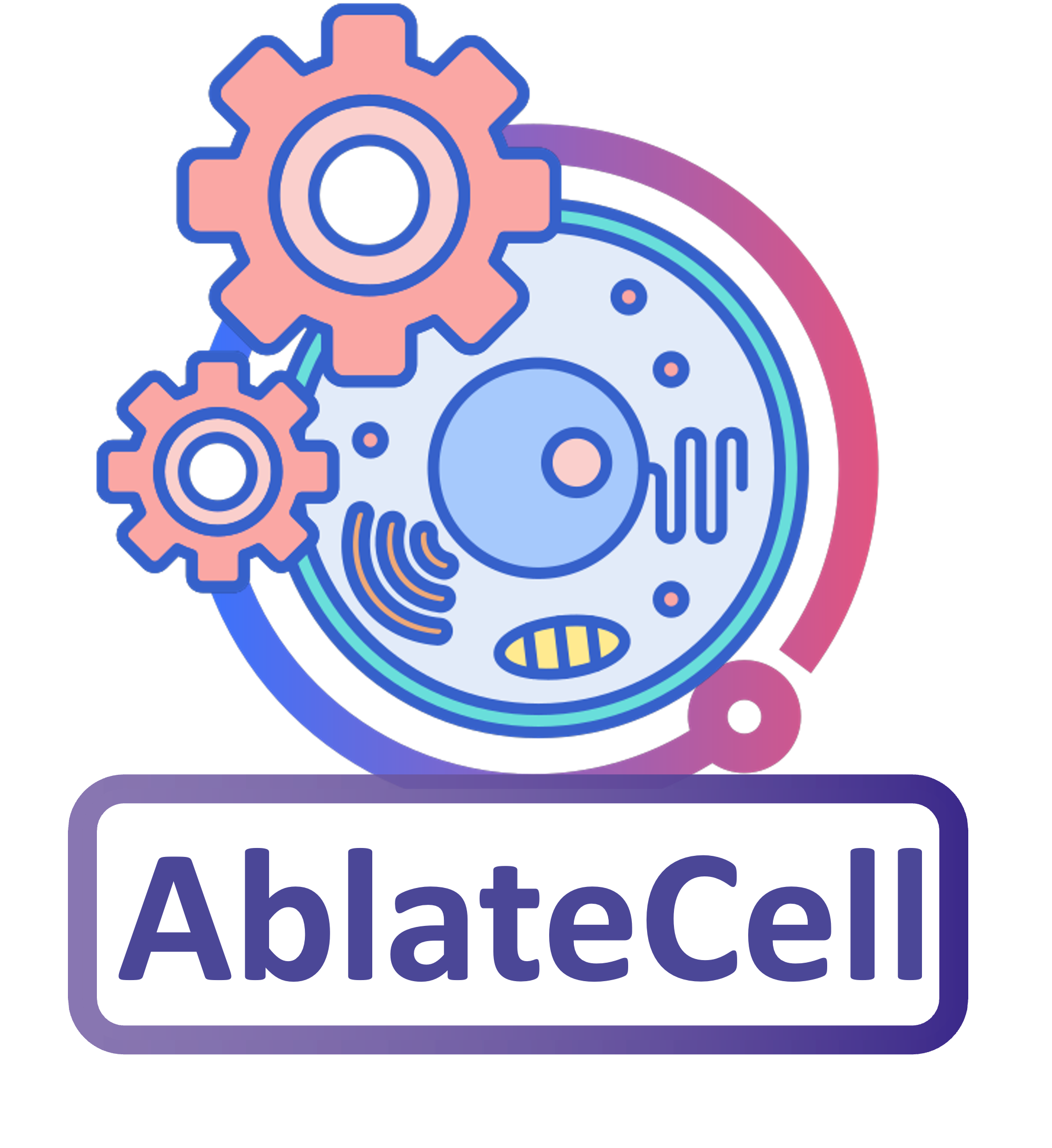}}
}

\newcommand{\eg}{\hbox{\emph{e.g.,}}\xspace}

\newcommand{\successrate}{88.9\%\xspace}
\newcommand{\componentacc}{93.3\%\xspace}

\icmltitlerunning{AblateCell: A Reproduce-then-Ablate Agent for Virtual Cell Repositories}

\begin{document}
\twocolumn[
  \icmltitle{\titleLogo \xspace \ours :  A Reproduce-then-Ablate Agent for Virtual Cell Repositories}
  
  \icmlsetsymbol{equal}{*}

  \begin{icmlauthorlist}
    \icmlauthor{Xue Xia}{ailab,hkustgz,equal}
    \icmlauthor{Chengkai Yao}{ucsd,equal}
    \icmlauthor{Mingyu Tsoi}{hkust}
    \icmlauthor{Xinjie Mao}{ailab,cz}
    \icmlauthor{Wenxuan Huang}{ailab,sch}
    \icmlauthor{Jiaqi Wei}{ailab}
    \icmlauthor{Hao Wu}{ailab,sch}
    \icmlauthor{Cheng Tan}{ailab}
    \icmlauthor{Lang Yu}{ailab}
    \icmlauthor{Yuejin Yang}{ailab,sch}
    \icmlauthor{Mengdi Liu}{ict}
    \icmlauthor{Siqi Sun}{ailab,sch}
    \icmlauthor{Zhangyang Gao}{ailab}
  \end{icmlauthorlist}

  \icmlaffiliation{hkustgz}{The Hong Kong University of Science and Technology (Guangzhou)}
  \icmlaffiliation{ucsd}{University of California San Diego}
  \icmlaffiliation{hkust}{The Hong Kong University of Science and Technology}
  \icmlaffiliation{ailab}{Shanghai Artificial Intelligence Laboratory}
  \icmlaffiliation{sch}{Fudan University}
  \icmlaffiliation{cz}{Shanghai Innovation Institute}
  \icmlaffiliation{ict}{Institute of Computing Technology, Chinese Academy of Sciences}  

  \icmlcorrespondingauthor{Siqi Sun}{siqisun@fudan.edu.cn}
  \icmlcorrespondingauthor{Zhangyang Gao}{gaozhangyang@ailab.org.cn}

  \vskip 0.3in
]



\printAffiliationsAndNotice{}  

\begin{abstract}
\input{main/0-abstract}

\end{abstract}

\input{main/1-introduction}

\input{main/2-related_work}
\input{main/3-method}
\input{main/4-experimental_steup}

\input{main/5-experiments}
\input{main/6-conclusion}

\section*{Impact Statement}

This paper presents work whose goal is to advance the field of machine learning by enabling automated ablation studies for computational biology models. 
While our system aims to democratize rigorous model validation and improve reproducibility in AI-driven biological research, we acknowledge that automated systems should augment rather than replace domain expertise and human judgment in scientific inquiry.
There are many potential societal consequences of our work, none of which we feel must be specifically highlighted here.

\bibliography{custom}
\bibliographystyle{icml2026}

\clearpage
\onecolumn
\appendix

\input{appendix/01_implementation_details}
\input{appendix/02_experimental_results}

\input{appendix/03_case_studies}
\input{appendix/04_prompts_and_templates}


\end{document}

%% file: main/0-abstract.tex

Systematic ablations are essential to attribute performance gains in AI Virtual Cells, yet they are rarely performed because biological repositories are under-standardized and tightly coupled to domain-specific data and formats. While recent coding agents can translate ideas into implementations, they typically stop at producing code and lack a verifier that can reproduce strong baselines and rigorously test which components truly matter. We introduce AblateCell, a \textbf{reproduce-then-ablate} agent for virtual cell repositories that closes this verification gap. AblateCell first reproduces reported baselines end-to-end by auto-configuring environments, resolving dependency and data issues, and rerunning official evaluations while emitting verifiable artifacts. It then conducts closed-loop ablation by generating a graph of isolated repository mutations and adaptively selecting experiments under a reward that trades off performance impact and execution cost. Evaluated on three single-cell perturbation prediction repositories (CPA, GEARS, BioLORD), AblateCell achieves \textbf{88.9\%} (+29.9\% to human expert) end-to-end workflow success and \textbf{93.3\%} (+53.3\% to heuristic) accuracy in recovering ground-truth critical components. These results enable scalable, repository-grounded verification and attribution directly on biological codebases.

%% file: main/1-introduction.tex
\section{Introduction}

Recently, the advent of large language models (LLMs) \cite{openai_gpt5_systemcard, anthropic_claude_sonnet_4_5_system_card, google_gemini_3_pro_model_card} has demonstrated remarkable capabilities in natural language understanding, reasoning, and code generation. Building upon these advances, researchers have developed LLM-based autonomous agents by incorporating planning, tool use, and memory, enabling them to address increasingly complex real-world tasks \cite{yao2023react, guo2024large, hu2025hiagent}.
These agents have demonstrated success across diverse domains, with recent advances enabling autonomous scientific discovery \cite{wei2025ai, hu2025survey}. 
In particular, multi-agent systems have focused on automating paper comprehension and code implementation, with representative works aiming to translate research ideas or method descriptions into executable programs \cite{lu2024ai, tian2024scicode, yamada2025ai}.
However, as generated research ideas and codes become increasingly complex, reliably reproducing, isolating and evaluating their effective components remains an open and fundamental challenge in specialized scientific domains \cite{heil2021reproducibility, van2024encore, zhou2025scientists, xu2025probing}.

\input{figures_tex/Figure1}


\textbf{Moving from idea synthesis to idea attribution in scientific domains is non-trivial: the full reproduce-ablate-analyze pipeline frequently breaks at multiple stages}, including reproduction, controlled repository edits, and domain-grounded reasoning. Firstly, the workflows may fail before reproduction, as MLR-Bench \cite{chen2025mlr} reports that roughly 80\% of automated experiments become invalid due to configuration and environment errors. Secondly, controlled ablation requires making precise, executable repository edits, while ResearchCodeBench \cite{hua2025researchcodebench} shows that even top LLMs succeed in implementing paper-level method changes in fewer than 40\% of cases. Thirdly, designing informative ablations and interpreting their outcomes requires substantial domain knowledge (e.g., mapping scientific hypotheses to code modules and metrics), which general-purpose coding agents are not optimized for. Consequently, scientific ablation is still largely manual \cite{fostiropoulos2023ablator} and frequently omitted in practice \cite{kargaran2025insights}. Considering these challenges, we \emph{re-scope the problem to a setting that matches realistic agent capabilities}: starting from a runnable baseline repository, we focus agents on controlled, isolated ablations--rather than end-to-end paper-to-executable re-implementation as in Paper2Agent~\citep{miao2025paper2agent}--to offload the code-learning burden from humans and enable scalable attribution.

In this paper, we introduce \ours, an end-to-end reproduce-then-ablate agent that turns scientific ablation from a brittle manual process into a \emph{verifiable} closed loop for AIVC repositories. \ours unifies paper grounding, baseline reproduction, and systematic ablation as a \textbf{graph-based multi-agent execution process} with explicit end-to-end coordination. It first establishes a verified baseline by auto-configuring environments, resolving dependency and data issues, and rerunning the official training and inference pipeline while emitting reproducible artifacts. It then performs closed-loop ablation via dynamic graph execution with adaptive sampling under a reward that trades off performance impact and execution cost, using isolated code mutation to prevent interference. We additionally construct a domain knowledge base by parsing AIVC papers and the original codebases, which grounds the ablation hypothesis proposal and supports agentic reasoning for evidence-based interpretation of ablation outcomes.

Comprehensive evaluation on CPA, GEARS, and BioLORD shows that \ours autonomously completes reproduce-then-ablation studies with \successrate success rate while identifying performance-critical components with \componentacc accuracy.
Specifically, \ours attains 96.3\% reproduction TSR (+26.9\% vs. RepoMaster, +24.0\% vs. human experts), 92.0\% ablation TSR (+12.0\% vs. human experts, +46.2\% vs. Mini-SWE-Agent), and 88.9\% end-to-end TSR (+29.9\% vs. human experts).
Our contributions are:

\begin{itemize}
    \item \textbf{\ours}: an end-to-end agentic verifier for AIVC single-cell perturbation repositories, achieving 96.3\% reproduction TSR and producing verifiable artifacts.
    \item A graph-based autonomous ablation engine with adaptive bandit sampling, isolated repo mutations, and domain-knowledge retrieval.
    \item Comprehensive experiments show that \ours significantly outperforms existing agent baselines, achieving 88.9\% end-to-end task success rate.
\end{itemize}

%% file: figures_tex/Figure1.tex
\begin{figure}[!t]
    \centering
    \includegraphics[width = \linewidth]{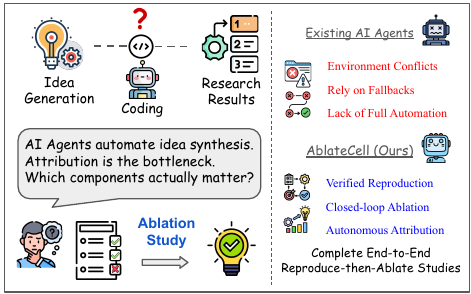}
    \caption{Motivation for \ours. Existing AI Agents scale idea synthesis but not idea attribution. Automated ablation bridges this gap by systematically verify which components truly matter.
    }
    \label{fig:task_overview}
\end{figure}

%% file: main/2-related_work.tex
\section{Related Work}

\input{figures_tex/overview}

\subsection{Code Generation for Scientific Workflows}
Large language models (LLMs) have shown strong capabilities in automated code generation~\citep{zhang2023repocoder, zhang2024codeagent, seo2025paper2code} and have been adapted to scientific workflows~\citep{tian2024scicode, nejjar2025llms}. 
Code-execution agents like SWE-agent and OpenHands incorporate repository interaction~\citep{yang2024swe, DBLP:conf/iclr/0001LSXTZPSLSTL25}, but they are not tailored for scientific reproduction and ablation, particularly in handling domain-specific data formats and experimental workflows~\citep{kuang2025process}.
Additionally, emerging benchmarks have been proposed for reproduction verification~\citep{siegel2025core, hu2025repro} and ablation planning~\citep{Zhao2025AbGen, abramovich2025ablationbench}, yet research-extension benchmarks report low end-to-end success rates~\citep{edwards2025rexbench}.



\subsection{Autonomous Research Agents}

Autonomous research agents combine reasoning, tool use, and iterative execution to automate scientific workflows and produce verifiable research outputs~\citep{zheng2025automation, gridach2025agentic, wei2025unifying}. 
Recent advancements range from knowledge-graph-driven multi-agent reasoning~\citep{ghafarollahi2025sciagents} to end-to-end research automation~\citep{yamada2025ai, baek2025researchagent}.
In single-cell biology, recent work has explored LLM agents for annotation~\citep{mao2025scagent}, image segmentation~\citep{yu2025gencellagent}, and perturbation prediction~\citep{tang2025cellforge}. 
Existing agents focus on forward research pipelines and lack systematic ablation capabilities~\citep{sheikholeslami2025utilizing}, motivating our reproduce-then-ablate framework that covers reproduction, hypothesis generation, and execution.

\subsection{Virtual Cell Models}

Virtual cell models learn mappings from unperturbed molecular profiles to perturbed outcomes~\citep{bunne2024build}. Existing perturbation predictors broadly fall into: (i) latent generative models (e.g., VAE-based) that encode cell state and perturbation effects for out-of-distribution generalization, often with disentangled factors for compositional inference~\citep{lotfollahi2019scgen, bereket2023samsvae, wang2024scLAMBDA, lotfollahi2023cpa, piran2024biolord}; (ii) GNN-based methods that inject gene-regulatory or interaction priors for mechanistic prediction~\citep{roohani2023gears}; and (iii) diffusion and transformer approaches that model complex state transitions or distributional shifts with large-scale pretraining~\citep{luo2024scdiffusion, he2025squidiff, liang2025scppdm, adduri2025state}. Accurate virtual cell models can reduce reliance on costly wet-lab experimentation and support therapeutic response prediction and target discovery~\citep{bunne2024build, ma2025aivcpotential}.

%% file: figures_tex/overview.tex
\begin{figure*}[!t]
    \centering
    \includegraphics[width = \linewidth]{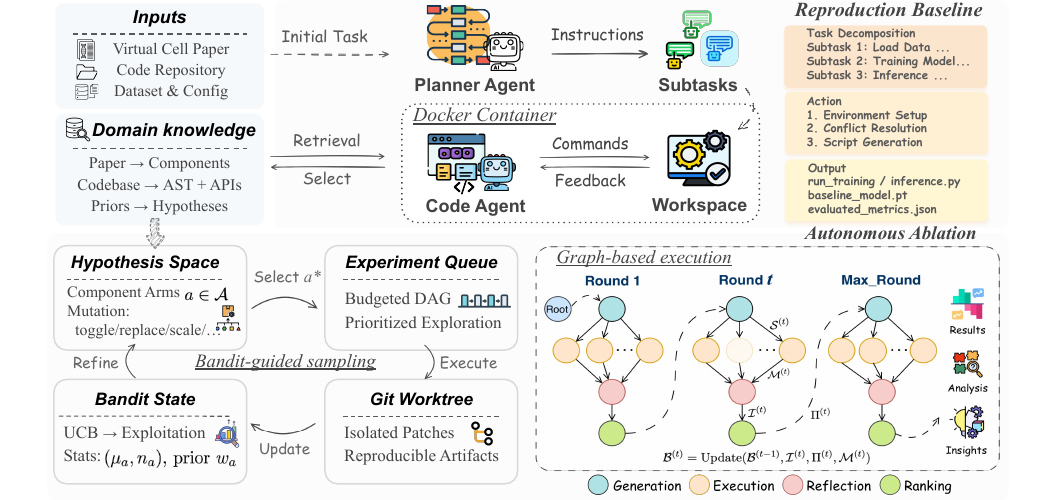}
    \caption{Overview of the \ours reproduce-then-ablate framework. The system (i) \textbf{reproduces baselines} via planner-executor agents in Docker container, then (ii) conducts \textbf{autonomous ablation} by selecting hypotheses with bandit sampling and executing via graph-based workflow in isolated worktrees, guided by domain knowledge throughout.}
    \label{fig:system_overview}
\end{figure*}

%% file: main/3-method.tex
\section{Method}
\label{sec:method}


We introduce \ours, an end-to-end autonomous reproduce-then-ablate multi-agent system for Virtual Cell repositories, as illustrated in Figure~\ref{fig:system_overview}.







\subsection{Problem Definition: Ablation as Information Gain}

Let $C=\{c_1,\dots,c_m\}$ denote the set of model components.
Let $f(\cdot)$ be a scalar performance score of the model.
For each component $c_i$, we define its (signed) ablation effect
\begin{equation}
\label{eq:component_importance}
\Delta_i \triangleq f(C) - f(C \setminus \{c_i\}),
\end{equation}
and measure its importance by $s_i \triangleq |\Delta_i|$. Components with $s_i \geq 0.05 \cdot |f(C)|$ (5\% relative change) are deemed critical.

\paragraph{Budgeted objective.}
Exhaustively evaluating all $m$ components is often infeasible. We therefore consider a budget of $B$ ablation runs ($B \ll m$) and aim to identify the top-$k$ most critical components:
\[
S_k^\star \triangleq \arg\max_{S\subseteq[m],\,|S|=k}\ \sum_{i\in S} s_i.
\]
Here, $S_k^\star$ represents the ground-truth top-$k$ components, validated by experts (Appendix~\ref{appendix:acc_k}), whose mutation causes the most significant performance change as measured by $s_i$.

\paragraph{Evaluation / ``information gain'' (intuitive).}
We interpret \emph{information gain} as reducing uncertainty about which components belong to $S_k^\star$.
Accordingly, we evaluate a policy by (i) top-$k$ recovery accuracy $\Pr(\hat S_k = S_k^\star)$, or equivalently
(ii) the expected simple regret:
\[
\mathcal{R} \triangleq \sum_{i\in S_k^\star} s_i - \sum_{i\in \hat S_k} s_i.
\]
This formulation naturally leads to bandit-style strategies that choose the next ablation to maximally shrink uncertainty
about large $s_i$ under a limited budget.

\subsection{Automated Baseline Reproduction}
\label{sec:baseline_reproduction}

Reproducing single-cell perturbation baselines is fragile due to complex dependencies and assumptions. \ours resolves this via planner-executor decomposition.

\paragraph{Planner Agent: High-level coordination.}
Given a reproduction task specification (\eg train the baseline, run inference), the planner agent decomposes it into concrete subtasks (data loading, training, inference, visualization) and constructs a grounded instruction for the code agent.
This instruction includes: (1) repository-specific API patterns extracted from training scripts, evaluation code, and configuration files; (2) required dataset paths and configuration keys; (3) expected outputs and execution constraints.
In parallel, the planner queries the domain knowledge base $\mathcal{K}$ (see Section~\ref{sec:domain_knowledge}) and performs agentic selection over retrieved entries, deciding what to include based on the task intent (\eg training vs.\ inference).
The planner operates in a containerized, agent-driven workflow environment to ensure reproducibility and consistency.

\paragraph{Code Agent: Low-level execution and debugging.}
The code agent receives the planner-generated instructions and performs deterministic execution inside a Docker container with the repository mounted at a stable path.
It focuses on implementation and debugging details: writing scripts, running commands, inspecting execution feedback (stdout/stderr and intermediate outputs), and iteratively editing code to resolve failures such as missing imports, dependency version mismatches, incorrect file paths, or data format incompatibilities.
This automated debugging loop continues iteratively until the baseline runs successfully or reaches the maximum iteration limit, enabling robust reproduction across different repository implementations and computational environments without requiring manual intervention.


\subsection{Graph-Based Ablation with Adaptive Sampling}
\label{sec:ablation_graph}

\paragraph{Notation.}
We define an ablation configuration \( x \in \mathcal{X} \) targeting a subset of components \( S(x) \subseteq C \). Arms are denoted by \( a \in \mathcal{A} \), where \( g(x) \in \mathcal{A} \) represents the arm corresponding to configuration \( x \). We perform experiments over \( T \) rounds, with a budget of \( B \) executed configurations.

Systematic ablation studies require orchestrating multiple interdependent experiments while efficiently exploring the exponentially large space of candidate ablations under computational constraints.
We formulate the ablation process as a directed acyclic graph (DAG), where nodes represent operations and edges encode dependencies between them.
To prioritize experiments, we employ a modified multi-armed bandit algorithm where each arm represents a candidate \emph{hypothesis family}.
The number of evaluations of each arm is denoted by \( n_a \), representing the number of times arm \( a \) has been selected. As \( n_a \) increases, the exploration bonus for arm \( a \) decreases, and the bandit algorithm increasingly relies on the empirical mean reward of the arm, thus favoring exploitation over exploration. 
This ensures less-explored arms are prioritized, balancing exploration and exploitation.

\input{algorithm/algorithm}

Additionally, we incorporate domain knowledge through weights \( w_a \), which prioritize certain arms for exploration. The weight \( w_a \) is based on prior knowledge or expert insights about the importance of certain components in the model. A higher \( w_a \) suggests that the corresponding arm is more likely to provide valuable information and should be explored more intensively. This modification enables scientifically-informed exploration, guiding the algorithm to explore more promising configurations first, rather than relying solely on random exploration or empirical data.
Our complete approach is detailed in Algorithm~\ref{alg:dynamic_ucb}.

The execution graph organizes each ablation round $t$ as a directed pipeline of dependent operation nodes.
A candidate \textbf{generation} node expands the selected arm $a^\star$ into executable configurations using heuristic generation rules defined over $\mathcal{X}$.
Each candidate then passes through an \textbf{execution} node run by a code agent, which trains the ablated model, runs inference, and outputs the ablation metrics $\mathcal{M}^{(r)}$.
After execution, \textbf{reflection} nodes aggregate and interpret the resulting metrics to produce insights $\mathcal{I}^{(r)}$, and \textbf{ranking} nodes compare candidates to produce rankings $\Pi^{(r)}$.
The selected results are used to update the bandit state $\mathcal{B}^{(r)}$ by incorporating new reward observations and updating empirical means $\mu_a$ and visit counts $n_a$, closing the loop and guiding candidate generation in the next round.
This graph-based representation makes dependencies explicit, enabling safe parallelism when operations are independent while preserving correct ordering and consistency.

\paragraph{Reward.}
We cast adaptive ablation as a bandit-driven hypothesis testing process: each arm proposes ablations, and the objective is to identify critical components under a finite compute budget. For each executed configuration \( x \), we define its ablation effect following Equation~\ref{eq:component_importance}:

\begin{equation}
\label{eq:config_ablation_effect}
\Delta(x) \triangleq f(C) - f(x),
\end{equation}

where $f(C)$ and $f(x)$ are the baseline and post-ablation scores using the primary metric.

We define the reward as:

\begin{equation}
\label{eq:reward}
r(x) \triangleq |\Delta(x)| - \lambda \cdot \mathrm{cost}(x),
\end{equation}

where $|\Delta(x)|$ quantifies the absolute performance impact, corresponding to component importance $s_i$ in Equation~\ref{eq:component_importance}. The term $\mathrm{cost}(x)$ quantifies computational cost in GPU-hours. The hyperparameter $\lambda$ balances informativeness against cost (see Appendices~\ref{appendix:system_configuration} and \ref{appendix:experimental_setup} for hyperparameter values and criticality threshold).

The reward $r(x)$ is attributed to the generating arm $g(x)$ to update bandit statistics, guiding subsequent exploration toward high-impact, cost-efficient ablations.



\subsection{Isolated Code Mutation with Git Worktree}
\label{sec:isolated_mutation}

Conducting multiple ablation experiments requires maintaining different code versions while ensuring each variant can be executed independently without conflicts.
For each candidate $x \in \mathcal{S}^{(r)}$, we instantiate an isolated Git worktree $w_{x_i}$ rooted at a fixed base commit.
Compared to cloning, worktrees share the same object database, enabling faster workspace creation and teardown while maintaining separate working directories, build artifacts, and untracked files.
Within each worktree, the code agent applies the mutation specification, executes training and inference, and preserves all modified files, generated artifacts, and execution logs.
This isolation prevents cross-contamination between concurrent code mutations (\eg disabling modules, swapping implementations, or altering configurations), ensuring that each ablation can be precisely reproduced and maintaining full experimental traceability.

\subsection{Domain Knowledge Base}
\label{sec:domain_knowledge}

To incorporate task-specific expertise into the ablation workflow, we construct a domain knowledge base $\mathcal{K}$ from AIVC papers and single-cell perturbation model codebases, including their documentation, architecture descriptions, and training protocols. 
We parse code repositories to extract and summarize individual functions, classes, and modules at the implementation level, while extracting key concepts from papers and documentation; all units are embedded to enable similarity-based retrieval~\citep{weiretrieval}.
Given a query $q$ derived from the paper or codebase (\eg extracted keywords such as "attention mechanism" or "cell-type stratification"), the retrieval function $\text{Retrieve}(q, \mathcal{K})$ returns the top-$k_{\text{ret}}$ most relevant component references $\{d_1,\dots,d_{k_{\text{ret}}}\}$ along with their descriptions to augment the agent’s context.

This retrieval process is integrated throughout both stages. During reproduction, it provides implementation guidance to the code agent. 
During ablation, the orchestrator uses retrieved knowledge to generate hypotheses and plan experiments, while the analyzer uses it for result interpretation and component ranking.
This domain knowledge tailors \ours to single-cell perturbation modeling.

%% file: algorithm/algorithm.tex
\begin{algorithm}[t]
\caption{Adaptive Ablation Study with Dynamic UCB}
\label{alg:dynamic_ucb}
\begin{algorithmic}[1]
\REQUIRE Max rounds $R$, candidate space $\mathcal{X}$, baseline score $f_b$, domain knowledge base $\mathcal{K}$
\ENSURE Best candidate and ablation insights

\STATE Initialize BanditState $\mathcal{B}$ with empty arms; $round \gets 0$; $T \gets 0$

\WHILE{$round < R$}
  \STATE \COMMENT{Dynamic exploration parameter}
  \STATE $\beta \gets \begin{cases} 1.5\beta_{\text{base}} & \text{if } round < 0.3R \\ 0.5\beta_{\text{base}} & \text{if } round \ge 0.7R \\ \beta_{\text{base}} & \text{otherwise} \end{cases}$
  
  \STATE \COMMENT{Arm selection}
  \STATE $unexplored \gets \{a \in \mathcal{B}.arms \mid n_a = 0\}$
  \IF{$unexplored \neq \emptyset$}
    \STATE $a^\star \gets \arg\max_{a \in unexplored} w_a$
  \ELSE
    \STATE $a^\star \gets \arg\max_{a}\Big[\mu_a + \beta \sqrt{\frac{\ln(T+1)}{n_a}}\Big]$
  \ENDIF
  
  \STATE \COMMENT{Generation budget}
  \STATE $K \gets \begin{cases} K_{\text{explore}} & \text{if } n_{a^\star} < 3 \\ K_{\text{exploit}} & \text{if } n_{a^\star} > 10 \\ K_{\text{base}} & \text{otherwise} \end{cases}$
  
  \STATE \COMMENT{Execute and update}
  \STATE $S \gets \textsc{GenerateCandidates}(\mathcal{X}, a^\star, K)$
  \FORALL{$x \in S$}
    \STATE $metrics \gets \textsc{Execute}(x)$; $r \gets \textsc{Reward}(f_b, metrics, \lambda, x)$
    \STATE $a \gets \textsc{ExtractArmID}(x)$; \textsc{UpdateBanditState}($\mathcal{B}, a, r$); $T \gets T + 1$
  \ENDFOR
  \STATE $round \gets round + 1$
\ENDWHILE
\end{algorithmic}
\end{algorithm}

%% file: main/4-experimental_steup.tex
\section{Experimental Setup}
\label{sec:experimental_setup}
In this section, we describe the experimental setting for evaluating \ours, including the target task and repositories, the comparison baselines, and the evaluation metrics.

\subsection{Task Description}
\label{sec:task_description}

We target single-cell perturbation prediction~\citep{bunne2024build}, which predicts gene expression changes in response to genetic perturbations (\eg gene knockout, overexpression, or knockdown).
Given baseline expression and a perturbation, models predict resulting changes across thousands of genes.
This task's controlled interventions enable systematic ablations, while its high dimensionality and reproducibility challenges demand autonomous ablation.


\paragraph{Target models.}
We evaluate \ours on three representative single-cell perturbation prediction models, each addressing distinct challenges in the field:

\begin{itemize}[leftmargin=*,nosep]
    \item \textbf{BioLORD}~\citep{piran2024biolord}: Disentangled representation learning framework that enables cross-condition transfer by separating biological and technical variations.
    \item \textbf{GEARS}~\citep{roohani2023gears}: Graph-based model that uses graph neural networks to capture combinatorial effects of multi-gene perturbations.
    \item \textbf{CPA}~\citep{lotfollahi2023cpa}: Compositional Perturbation Autoencoder that generalizes to unseen perturbation combinations and cell types through disentangled compositional representations.
\end{itemize}

These models span different architectural paradigms (autoencoders, graph neural networks, disentangled representations) and modeling objectives (combinatorial generalization, cross-condition transfer), operating on heterogeneous single-cell datasets across different perturbation modalities and biological conditions (see Table~\ref{tab:dataset_statistics}, Appendix~\ref{appendix:dataset_statistics}), allowing us to assess whether \ours can autonomously navigate diverse modeling approaches and identify critical components across varied experimental settings.

\paragraph{Evaluation metrics for perturbation prediction.}
Following standard practice in single-cell perturbation prediction, we evaluate model performance mainly using two complementary metrics.
Let $\hat{y}_i$ and $y_i$ denote the predicted and true expression values for gene $i$ over $n$ genes:

\begin{align*}
&\text{MSE} = \frac{1}{n}\sum_{i=1}^{n}(\hat{y}_i - y_i)^2, \\
&\text{Pearson} = \frac{\text{Cov}(\hat{y}, y)}{\sigma_{\hat{y}} \sigma_{y}}.
\end{align*}

where $\text{Cov}(\cdot,\cdot)$ denotes covariance and $\sigma$ denotes standard deviation.
These metrics measure baseline performance and ablation impact in our experiments (Section~\ref{sec:experiments}).

\subsection{Baseline Methods}
\label{sec:baseline_methods}


We target the novel task of end-to-end reproduce-then-ablate studies, covering paper comprehension, script generation, experiment execution, and result analysis. While no existing methods address the full pipeline, several agent frameworks handle individual stages. We establish two baselines:

\begin{itemize}[leftmargin=*,noitemsep,topsep=4pt]
    \item \textbf{Human Performance Baseline}: Six experts (three computational biology researchers and three machine learning PhD students) independently reproduce baseline methods and design ablation experiments under closed-book conditions, with a time limit of 2 hours per task and a maximum of 5 debug attempts per experiment. The results are compared to expert consensus to quantify the gap between automated and human-guided studies.
    
    \item \textbf{Agent Framework Baselines}: Mini-SWE-Agent~\cite{yang2024swe} and RepoMaster~\cite{wang2025repomaster}, which cover key stages of the reproduce-then-ablate pipeline. Both of them perform one-shot execution for each task, with the allowance for self-debugging during the process.
\end{itemize}

\subsection{Evaluation Metrics}
\label{sec:evaluation}


We measure the performance of \ours through hypothesis quality, task completion rates, and resource efficiency.

\underline{\textit{Ablation hypothesis quality.}}
We assess generated ablation hypotheses through three dimensions:
(1) \textbf{Semantic coherence}: fraction of hypotheses that target meaningful architectural components rather than auxiliary code (e.g., logging, visualization):
\begin{equation}
\text{Sem.}=\frac{1}{N}\sum_{j=1}^{N}\mathbb{I}[\sigma_j=1],
\end{equation}
where $\sigma_j\in\{0,1\}$ indicates whether the $j$-th hypothesis targets a semantically meaningful component, determined by expert annotation;
(2) \textbf{Executability}: fraction of hypotheses returning valid metrics:
\begin{equation}
\text{Exec.}=\frac{1}{N}\sum_{j=1}^{N}\mathbb{I}[v_j=1],
\end{equation}
where $v_j\in\{0,1\}$ indicates whether the $j$-th ablation run returns valid evaluation metrics;
(3) \textbf{Component identification accuracy}:
\begin{equation}
\text{Acc@}k = \frac{|\text{Top-}k_{\text{pred}} \cap \text{Top-}k_{\text{gt}}|}{k},
\end{equation}
where \( \text{Top-}k_{\text{gt}} \) is the set of the \( k \) most critical components selected based on expert consensus (Appendix~\ref{appendix:acc_k}).


\underline{\textit{Task success rate.}}
To quantify execution capability across the reproduction and ablation stages, we define
\begin{equation}
\mathrm{TSR}=\frac{1}{N}\sum_{j=1}^{N}\mathbb{I}[\delta_j=1],
\end{equation}
where $\delta_j\in\{0,1\}$ indicates whether the $j$-th task completes without manual intervention and $N$ is the number of tasks.
The end-to-end success rate is computed as the product $\text{TSR}_{\text{reproduction}} \times \text{TSR}_{\text{ablation}}$.


\underline{\textit{Operational efficiency.}}
We measure efficiency by API cost, median iterations per task, and wall-clock time to success. 
Time to success excludes model training and inference time to isolate the system's planning and execution overhead.

%% file: main/5-experiments.tex
\section{Experiments}
\label{sec:experiments}
We evaluate \ours on BioLORD, GEARS, and CPA to answer the following research questions:
\begin{itemize}[leftmargin=*,noitemsep,topsep=4pt]
    \item \textbf{RQ1}: Can \ours reliably execute end-to-end ablation workflows across the target codebases?
    \item \textbf{RQ2}: Does \ours generate high-quality hypotheses targeting performance-critical components?
    \item \textbf{RQ3}: What architectural components does \ours identify as critical for each target model?
\end{itemize}

\subsection{Main Results}
\label{sec:main_results}

\input{tables/comparison_table}

\paragraph{RQ1: Superior end-to-end ablation-to-insight performance.}
Table~\ref{tab:main_results_tsr} reports task success rates (TSR) across reproduction and ablation stages.
\ours achieves consistently high end-to-end TSR by seamlessly integrating all stages: the planner--executor decomposition reliably identifies correct entry points and resolves dependencies during reproduction, graph-based orchestration robustly handles diverse code mutations during ablation, and isolated worktree execution ensures clean experimental separation throughout the entire workflow.
By contrast, baseline agent frameworks often break due to (i) task-understanding drift, (ii) excessive reliance on generic fallbacks, (iii) execution hallucinations such as invoking nonexistent files, entry points, or CLI arguments, and (iv) inability to correctly handle domain-specific data formats such as h5ad files used in single-cell genomics, which lead to cascading failures without effective recovery.

\paragraph{RQ2: Effective hypothesis generation and validation.} 
Table~\ref{tab:hypothesis_quality} evaluates ablation hypothesis quality across semantic coherence, executability, and component identification accuracy. 
\ours generates high-quality hypotheses with perfect semantic coherence and high executability across both LLM backends, substantially outperforming random sampling (42.8\%, 35.7\%) and rule-based heuristic methods (68.5\%, 64.3\%).
\ours achieves 93.3\% Acc@5 in identifying performance-critical components, where \( k = 5 \) was chosen as it strikes a balance between model complexity and interpretability, focusing on the most critical components for performance. This demonstrates that the bandit-guided search effectively prioritizes impactful architectural elements over uninformed exploration strategies.

\input{tables/hypothesis_quality}

\paragraph{RQ3: Identification of performance-critical components.}
Figure~\ref{fig:component_impact} shows the key architectural components identified by \ours across three tasks (detailed results in Appendix \ref{appendix:component_impact}).
The analysis reveals task-specific critical elements: for CPA, the unified latent embedding shows the highest impact (30.35\% MSE increase upon removal), while the adversarial discriminator exhibits minimal impact (2.86\% MSE decrease upon ablation).
In GEARS, the perturbation GNN encoder emerges as overwhelmingly critical (89.70\% MSE increase), significantly outweighing other components.
For BioLORD, the unknown-attribute embedding ($z_u$) proves most impactful (63.05\% MSE increase), validating its role in capturing biological variability.
These findings are consistent with prior work~\citep{wu2024perturbench}, validating that \ours can automatically recover expert-level insights and provide actionable guidance: prioritize high-impact components for refinement and remove low-impact elements for simplification. 
Our results suggest that \ours not only helps identify critical components but also provides actionable strategies for model optimization.

\input{figures_tex/component_importance}

\subsection{Ablation Study}
\label{sec:ablation_studies}

To understand the contribution of each component in \ours, we perform a systematic ablation study by removing key modules and measuring their impact on task success rate (TSR), component identification accuracy (Acc@5), and API cost.
Table~\ref{tab:system_ablation} summarizes the results.

Removing the \textbf{planner-executor decomposition} leads to the most substantial TSR drop (88.9\% $\rightarrow$ 65.4\%), as the unified agent struggles with context management and fails to maintain a clear separation between planning and execution, while API costs increase by 55\% due to repeated trial-and-error without structured planning.
Ablating \textbf{domain knowledge retrieval} significantly impacts both execution reliability and hypothesis quality, degrading TSR by 18.7\% and Acc@5 by 23.3\%, as the system lacks domain-specific understanding of component semantics and must rely on blind exploration—leading to more invalid hypotheses and wasted trials.
Replacing \textbf{adaptive bandit sampling} with uniform random selection severely degrades Acc@5 (93.3\% $\rightarrow$ 53.3\%), demonstrating the critical importance of guided exploration; however, TSR remains nearly unchanged ($-$0.8\%), indicating that execution reliability is decoupled from hypothesis quality.
Without \textbf{git worktree isolation}, execution time increases dramatically (45.2 min $\rightarrow$ 78.5 min) due to the inability to parallelize ablation experiments and the overhead of manual state cleanup between trials; though TSR and Acc@5 remain unchanged, as the worktree is purely an efficiency mechanism.

These results reveal distinct roles for each component: planner-executor decomposition ensures execution reliability, domain knowledge bridges execution and hypothesis quality, git worktree isolation enables parallelization efficiency, and adaptive bandit sampling drives hypothesis quality.
The combination of all components achieves optimal balance across execution, insight generation, and efficiency.

\input{tables/ablation_study}

\subsection{Case Study}
\label{sec:case_studies}
We demonstrate \ours on GEARS, a graph neural network model designed to predict transcriptional responses to genetic perturbations, as illustrated in Figure~\ref{fig:example}. In this case study, \ours autonomously extracted the core components from the associated paper, generated targeted ablation hypotheses, and applied the UCB bandit algorithm with domain knowledge weighting to prioritize the exploration of the most impactful components.

The bandit algorithm identified the perturbation GNN encoder as a key component. Ablation experiments confirmed its critical importance: removing message passing on perturbation embeddings resulted in substantial performance degradation (89.7\% MSE increase, 0.012 Pearson drop). This significant deterioration demonstrates that perturbation-level message passing is essential for capturing complex interactions between genetic perturbations and their downstream effects, validating this architectural design choice.

Throughout the experiment, \ours showcased its core capabilities: automatic component extraction, intelligent hypothesis generation, and full automation of code modification, training, and evaluation. The UCB bandit algorithm, guided by domain knowledge, effectively steered exploration toward impactful components. The perturbation GNN encoder (5 trials) proved most critical, yielding the largest performance impact when ablated. This demonstrates \ours' efficiency in identifying critical components and quantitatively validating architectural design decisions through systematic experimentation.

\input{figures_tex/example}

%% file: tables/comparison_table.tex
\begin{table}[t]
\centering
\caption{Task success rate (\%) comparison across reproduction and ablation stages. Best results in each stage are shown in bold. \ours results are highlighted.}
\label{tab:main_results_tsr}
\setlength{\tabcolsep}{6pt}
\resizebox{\columnwidth}{!}{
\begin{tabular}{lcccc}
\toprule
\textbf{Method} & \textbf{BioLORD} & \textbf{GEARS} & \textbf{CPA} & \textbf{Avg.} \\
\midrule
\multicolumn{5}{l}{\textit{Reproduction Stage}} \\
\midrule
Human Experts   & 93.3 & 73.7 & 50.0 & 72.3 \\
Mini-SWE-Agent  & 0.0 & 25.0 & 22.2 & 15.7 \\
RepoMaster      & 100.0 & 75.0 & 33.3 & 69.4 \\
\rowcolor{blue!5}
\ours           & \textbf{100.0} & \textbf{100.0} & \textbf{88.9} & \textbf{96.3} \\
\midrule
\multicolumn{5}{l}{\textit{Ablation Stage}} \\
\midrule
Human Experts   & 86.2 & 83.3 & 70.6 & 80.0 \\
Mini-SWE-Agent  & 12.5 & 50.0 & 75.0 & 45.8 \\
RepoMaster      & 66.7 & 33.0 & 25.0 & 41.6 \\
\rowcolor{blue!5}
\ours           & \textbf{92.0} & \textbf{100.0} & \textbf{84.0} & \textbf{92.0} \\
\midrule
\multicolumn{5}{l}{\textit{End-to-End}} \\
\midrule
Human Experts   & 80.4 & 61.4 & 35.3 & 59.0 \\
Mini-SWE-Agent  & 0.0 & 12.5 & 16.7 & 9.7 \\
RepoMaster      & 66.7 & 25.0 & 8.3 & 33.3 \\
\rowcolor{blue!5}
\ours           & \textbf{92.0} & \textbf{100.0} & \textbf{74.8} & \textbf{88.9} \\
\bottomrule
\end{tabular}
}
\end{table}

%% file: tables/hypothesis_quality.tex
\begin{table}[t]
\centering
\caption{Ablation hypothesis quality averaged across BioLORD, GEARS, and CPA. 
All metrics are reported in percentages (\%).}
\label{tab:hypothesis_quality}
\setlength{\tabcolsep}{6pt}
\renewcommand{\arraystretch}{1.05}
\resizebox{\columnwidth}{!}{
\begin{tabular}{lccc}
\toprule
\textbf{Method} & \textbf{Sem. $\uparrow$} & \textbf{Exec. $\uparrow$} & \textbf{Acc@5 $\uparrow$} \\
\midrule
Random selection  & 42.8 & 35.7 & 13.3 \\
Heuristic-based   & 68.5 & 64.3 & 40.0 \\
\midrule
\rowcolor{blue!5}
\ours~(GPT-4o-mini)         & \textbf{100.0} & 92.0 & 86.7 \\
\rowcolor{blue!5}
\ours~(Claude-Sonnet-4.5)   & \textbf{100.0} & \textbf{94.8} & \textbf{93.3} \\
\bottomrule
\end{tabular}
}
\end{table}

%% file: figures_tex/component_importance.tex
\begin{figure}[t]
\centering
\includegraphics[width=\columnwidth, height=0.13\textheight]{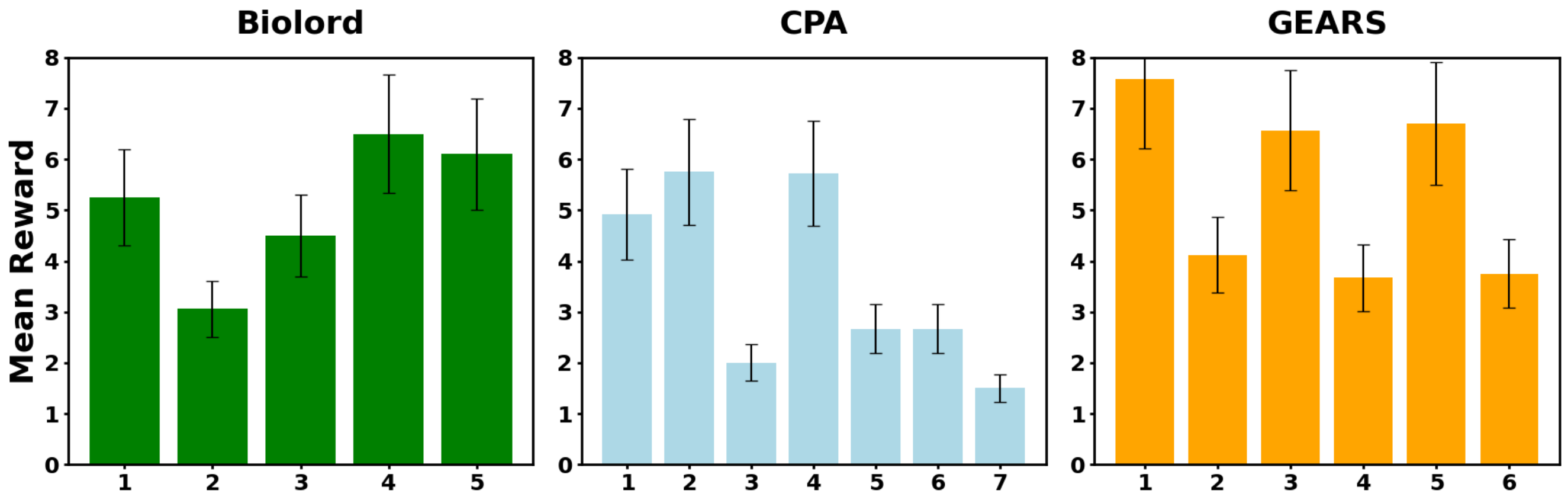}
\caption{Component importance across BioLORD, CPA, and GEARS. The x-axis shows component indices (names omitted), and the y-axis shows mean reward.}
\label{fig:component_impact}
\end{figure}

%% file: tables/ablation_study.tex
\begin{table}[t]
\centering
\caption{System ablation of \ours averaged across BioLORD, GEARS, and CPA. 
TSR and Acc@5 are reported in percentages (\%), Cost in dollars (\$), and Time in minutes. 
Relative changes with respect to the full system are shown in parentheses.}
\label{tab:system_ablation}
\setlength{\tabcolsep}{3pt}
\renewcommand{\arraystretch}{1.25}
\resizebox{\columnwidth}{!}{%
\begin{tabular}{lllll}
\toprule
\textbf{Configuration} & \textbf{TSR $\uparrow$} & \textbf{Acc@5 $\uparrow$} & \textbf{Cost $\downarrow$} & \textbf{Time $\downarrow$} \\
\midrule
\ours~(default) & 88.9 & 93.3 & 6.3 & 45.2 \\
\midrule
w/o Planner-executor   & 65.4~{\color{red}($-$23.5)} & 89.7~($-$3.6) & 9.8~{\color{red}(+3.5)} & 52.3~{\color{red}(+7.1)} \\
w/o Domain knowledge   & 70.2~{\color{red}($-$18.7)} & 70.0~{\color{red}($-$23.3)} & 8.1~{\color{red}(+1.8)} & 58.4~{\color{red}(+13.2)} \\
w/o Adaptive bandit    & 88.1~($-$0.8) & 53.3~{\color{red}($-$40.0)} & 6.7~(+0.4) & 48.3~(+2.1) \\
w/o Git Worktree       & 88.9~(0.0) & 93.3~(0.0) & 6.4~(+0.1) & 78.5~{\color{red}(+33.3)} \\
\bottomrule
\end{tabular}%
}
\end{table}

%% file: figures_tex/example.tex
\begin{figure}[t]
    \centering
    \includegraphics[width = 0.93\linewidth]{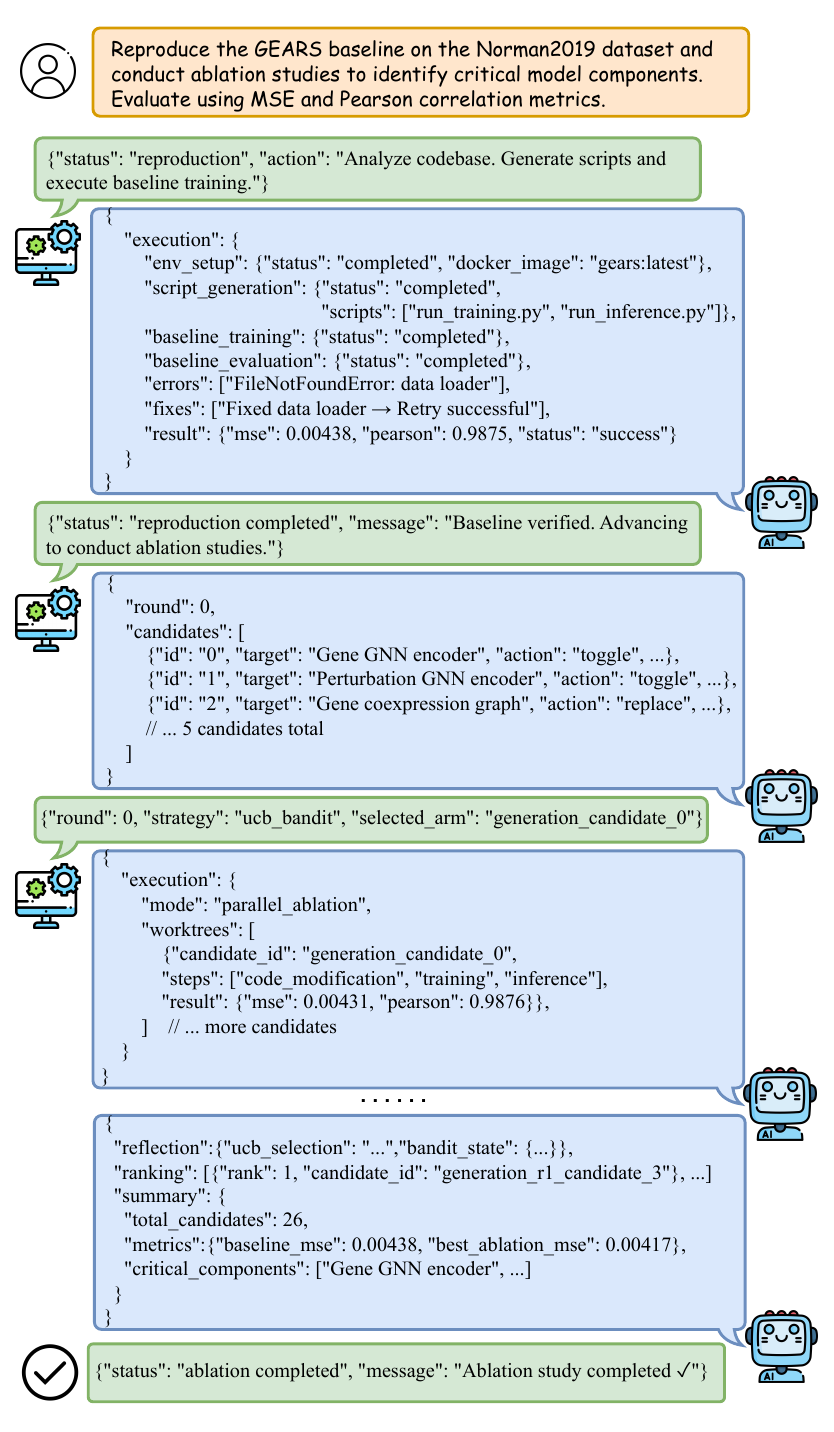}
    \caption{An example of \ours applied to GEARS for end-to-end reproduce-then-ablate execution.}
    \label{fig:example}
\end{figure}

%% file: main/6-conclusion.tex
\section{Conclusion}

We introduced \ours, a reproduce-then-ablate agent that enables systematic, automated ablation studies directly on single-cell perturbation prediction repositories.
By first reproducing reported baselines end-to-end through environment auto-configuration and dependency resolution, then conducting closed-loop ablation via graph-based mutation generation and adaptive bandit exploration, \ours achieves 88.9\% end-to-end workflow success and 93.3\% accuracy in identifying performance-critical components across CPA, GEARS, and BioLORD.
\ours closes the verification gap in AI-driven scientific workflows by autonomously discovering critical architectural elements that align with prior manual studies while providing actionable attribution insights for model refinement.

\textbf{Limitations.} \ours currently focuses on single-cell perturbation prediction with predefined domain knowledge; generalizing to diverse biological tasks with minimal priors is an important direction for future work.

%% file: appendix/01_implementation_details.tex
\section{Implementation Details}
\label{appendix:implementation}

\subsection{Domain Knowledge Base Construction}
\label{appendix:domain_knowledge}

In this section, we describe the \textbf{automated process} of constructing our domain knowledge base, designed to enhance the understanding and utilization of the target repositories.
The process is divided into two key stages: \emph{Python File Parsing} and \emph{Semantic Enrichment}.

\textbf{Python File Parsing with AST: }  In the first stage, we use the AST module to extract all methods and functions (both inside and outside classes) from python files in the target repositories. The extracted information, including functiom names, parameters, docstrings, and code snippets.
    
\textbf{Semantic Enrichment with LLM:}
In the second stage, we utilize LLM (gpt-4o) to generate concise descriptions for each functiom based on the extracted information. Using customized prompts for functions and class methods, the LLM produces descriptions that summarize the purpose and functionality of each element. These descriptions serve as an index for embedding knowledge into the agent workflow. 

The final output of this stage is a refined JSON structure, which combines the extracted code snippets with their corresponding semantic descriptions in the paper.

Importantly, the knowledge base is continuously expanded and updated based on experimental results. As new components are evaluated and new insights are gathered from the ablation studies, the knowledge base is automatically enriched with updated descriptions and additional components. This ensures that the knowledge base evolves in tandem with ongoing experiments, making it adaptable and scalable.

This approach is not limited to a specific domain; it is designed to be completely reusable for high-quality repositories from well-documented papers across various domains. As long as the repository maintains a high standard of code and documentation, the system can be applied to different types of research, ensuring broad applicability beyond just the initial case studies.

\subsection{Detailed Description of Candidate Space and Generation Rules}
\label{appendix:candidate_space_details}

This section provides a detailed description of the candidate space \( \mathcal{X} \) and the generation rules used to explore candidate configurations in the ablation study.

\subsubsection{Candidate Space \( \mathcal{X} \)}
The candidate space \( \mathcal{X} \) consists of all possible configurations derived from the components of the system. Each configuration represents a possible state of the system after applying a set of ablations to specific components. We define the candidate space as:

\[
\mathcal{X} = \{ x_1, x_2, \dots, x_N \}
\]

Where \( N \) is the total number of candidate configurations. Each configuration \( x_i \in \mathcal{X} \) is characterized by a set of ablations applied to the components \( C = \{ c_1, c_2, \dots, c_m \} \), the building blocks of the system. Each configuration is a subset of the components in \( C \).

\subsubsection{Generation Rules}
The process of generating candidate configurations is guided by several rules determining how ablations are applied. These rules combine heuristic methods, domain knowledge, and exploration strategies. The key steps involved are as follows:

\begin{itemize}
    \item \textbf{Heuristic Rule:} A heuristic rule determines which components are likely critical to the system's performance. For example, we prioritize components that are frequently modified in the literature or identified as key drivers in prior studies.
    
    \item \textbf{Exploration Strategy:} We employ an exploration strategy, such as Upper Confidence Bound (UCB), to select configurations that maximize information gain. This approach prioritizes configurations that have not been extensively explored, aiming to discover new insights.
    
    \item \textbf{Domain Knowledge-Based Generation:} Domain-specific knowledge is used to refine the generation process. Certain ablations may be restricted based on known interactions between components or expert recommendations, ensuring that the generated configurations are relevant to the scientific context.
    
    \item \textbf{Mutation Types:} Each configuration \( x_i \) can be modified through various mutations, such as toggling the presence of a component, replacing one component with another, or adjusting component hyperparameters. These mutations define the search space within \( \mathcal{X} \).
\end{itemize}

\subsubsection{Example of Candidate Generation}
Consider a system with three components \( C = \{ c_1, c_2, c_3 \} \). Example candidate configurations include:
- \( x_1 = \{ c_1 \} \) (removing \( c_1 \)),
- \( x_2 = \{ c_2, c_3 \} \) (removing \( c_1 \) and combining \( c_2 \) and \( c_3 \)),
- \( x_3 = \{ c_1, c_2 \} \) (removing \( c_3 \)).

The generation rules decide which components to ablate based on heuristics and exploration strategies.

\subsubsection{Impact of Generation Rules on Candidate Selection}
The generation rules significantly affect the efficiency and effectiveness of the exploration process. By prioritizing certain configurations, we aim to maximize the discovery of critical components while minimizing computational costs. The application of domain knowledge ensures that generated configurations are realistic and relevant to the task.

\subsection{Acc@k Definition and Ground Truth Validation}
\label{appendix:acc_k}

\paragraph{Acc@k Metric Definition.}  
Acc@k measures the accuracy of identifying the top-k most critical components in an ablation study. Specifically, Acc@5 calculates the proportion of correctly identified components within the top 5 predictions:
\begin{equation}
\text{Acc@k} = \frac{|\text{Predicted Top-k} \cap \text{Ground Truth Top-k}|}{k}
\end{equation}
where predicted components are ranked by performance impact (reward difference) and ground truth is established through expert consensus.

\paragraph{Expert Annotation Protocol.}
Ground truth was established by \textbf{3 independent domain experts per repository}, all with Ph.D. degrees and 2+ years of research experience in computational biology and machine learning. Experts independently reviewed ablation results and ranked components by criticality, following standardized guidelines that defined critical components as those fundamentally altering model functionality (not merely degrading performance metrics). The annotation process considered both empirical performance impact and theoretical importance grounded in domain knowledge. Components were deemed critical if their ablation resulted in $s_i = |\Delta_i| \geq 0.05 \cdot |f(C)|$ (5\% relative performance change).

\paragraph{Inter-Annotator Agreement.}
To ensure reliability and minimize subjectivity, we computed inter-annotator agreement using Fleiss' kappa ($\kappa$):
\begin{itemize}
    \item \textbf{Component criticality ranking}: $\kappa = 0.72$ (substantial agreement according to Landis \& Koch interpretation)
    \item \textbf{Hypothesis meaningfulness}: $\kappa = 0.68$ (substantial agreement)
    \item \textbf{Top-5 component overlap}: Average pairwise agreement = 82.7\%
\end{itemize}

\paragraph{Bias Mitigation.}
To prevent circular bias, annotators were not involved in designing the \ours system or constructing the domain knowledge base. Experts were blinded to which system generated each hypothesis and performed all annotations after experiments were completed. This independence ensures that ground truth reflects objective expert judgment rather than alignment with system design choices, contributing to the reliability and credibility of the Acc@k metric used in our evaluation.

%% file: appendix/02_experimental_results.tex
\clearpage
\section{Experimental Setup}

\subsection{Docker Environment Configuration}
\label{appendix:docker_config}

To ensure reproducibility and isolation across all experiments, we use Docker containers throughout the entire pipeline. Each model repository has a dedicated Dockerfile that defines a complete, isolated environment with all necessary dependencies for both reproduction and ablation stages.

All Docker images are based on \texttt{nvidia/cuda:12.1.0-cudnn8-runtime-ubuntu22.04}, providing CUDA 12.1 and cuDNN 8 support for GPU-accelerated training. The base configuration uses Ubuntu 22.04 LTS as the operating system, with CUDA runtime 12.1.0 and cuDNN 8. Python environments are managed via Miniconda with Python 3.10, and essential system dependencies including curl, git, and bash are pre-installed.

To be compatible with the system, Docker containers must: (1) expose a bash-compatible shell interface, (2) support GPU access when required (via NVIDIA Docker runtime), (3) maintain the standard directory structure, and (4) execute Python scripts in the configured Conda environment. The container's internal implementation (specific dependencies, library versions, or installation methods) remains opaque to the system, as long as these interface requirements are met.

This interface abstraction allows the Code Agent to execute training and inference scripts uniformly across different tasks without modification, while each task can maintain its own optimized environment configuration internally.

\subsection{Hyperparameters and System Settings}
\label{appendix:hyperparameters}

This section lists the hyperparameter settings and system configurations used in our experiments, covering all critical parameter settings, optimization algorithms, and training configurations to allow for the reproducibility of our experiments. We ensure that these configurations are consistent across all tasks (BioLORD, CPA, and GEARS) to facilitate reproducibility. 

\subsubsection{System Configuration}
\label{appendix:system_configuration}
Table~\ref{tab:common_hyperparams} provides a comprehensive summary of all hyperparameter configurations, training settings, evaluation configurations, and system settings that are common across all three tasks (BioLORD, CPA, and GEARS). These settings ensure reproducibility and consistency across our experimental framework.

\input{appendix/tables/system_hyperparameters}

Table~\ref{tab:task_parameters} details the task-specific training and configuration parameters that we explicitly set for each of the three tasks. These parameters include training configurations (epochs, batch size, random seed), data characteristics (number of cells, genes, data splits), and baseline performance metrics. All parameters not explicitly listed follow the default configurations from the respective model repositories.

\input{appendix/tables/task_parameter}

\paragraph{Baseline Reproduction and Metric Validation.}
To strengthen the distinction between execution success (TSR) and reproduction correctness, we validate that our reproduced baselines match reported metrics from the original papers. Our reproduction protocol strictly follows the default configurations provided in the authors' open-source repositories without additional hyperparameter tuning.

We acknowledge that published results in papers may reflect carefully tuned hyperparameters or multiple experimental runs selected for optimal performance, while repository defaults may represent more general configurations. Therefore, we adopt a tolerance-based validation approach:

\begin{itemize}
    \item \textbf{Tolerance threshold}: $\pm 5\%$ relative error for primary metrics (MSE, Pearson correlation, retrieval accuracy).
    \item \textbf{Common sources of deviation}: Random seed differences, hardware-specific numerical precision, and unspecified hyperparameters in papers.
\end{itemize}

This validation approach ensures that our "verified reproduction" claim is based on both autonomous execution (TSR) and scientific correctness, while acknowledging the inherent variability between paper-reported results and repository defaults.

\clearpage
\section{Experiments}
\label{appendix:appendix_experiments}

\subsection{Dataset Statistics}
\label{appendix:dataset_statistics}

We evaluate \ours on three biological model reproduction tasks, each associated with a
public single-cell or gene-expression dataset that we follow from the original papers.

\paragraph{CPA (single-cell perturbation prediction).}
For CPA, we follow the original CPA paper and use the Norman2019 CRISPR perturbation
single-cell RNA-seq dataset~\citep{norman2019}. The model predicts gene expression
responses under genetic or drug perturbations. In our evaluation split, the test set
contains $16{,}669$ cells and $5{,}044$ highly variable genes (HVGs), corresponding to
the processed file \texttt{Norman2019\_normalized\_hvg.h5ad} released by the CPA authors,
yielding a prediction tensor of shape $16{,}669 \times 5{,}044$.

\paragraph{GEARS (graph-based perturbation prediction).}
For GEARS, we follow the official GEARS implementation and use the same Norman CRISPR
perturbation dataset as in the original work~\citep{roohani2023gears}, obtained from Dataverse
as described in the GEARS repository. The graph neural network predicts transcriptomic
responses for combinatorial perturbations. In our setting, the evaluation split comprises
$28{,}754$ test cells and $5{,}045$ genes, yielding a prediction tensor of shape
$28{,}754 \times 5{,}045$.

\paragraph{BioLORD (latent representation learning for omics).}
For BioLORD, we follow the official tutorials and use the single-cell infection dataset
distributed with the Biolord package (files \texttt{adata\_abortive.h5ad} and
\texttt{adata\_infected.h5ad}) from Piran et al.~\citep{piran2024biolord}. The
task focuses on learning latent representations that disentangle known and unknown
attributes across different infection states. The evaluation split used in our experiments
contains $2{,}859$ cells and $8{,}203$ genes, corresponding to a prediction tensor of shape
$2{,}859 \times 8{,}203$.

\input{appendix/tables/dataset_statistics}


\subsection{Experimental Setup}
\label{appendix:experimental_setup}
We configure \ours with the following settings:

\begin{itemize}[topsep=3pt, itemsep=2pt, parsep=1pt]
    \item \textbf{UCB coefficient}: 2.0 (with adaptive adjustment based on exploration phase)
    \item \textbf{Max candidates per round}: 5
    \item \textbf{Max rounds}: 5
    \item \textbf{Reward metric}: Pearson correlation coefficient (primary), with MSE and MAE as secondary metrics
    \item \textbf{Cost penalty}: $\lambda$ = 0.01 (cost penalty coefficient for balancing performance and computational cost)
    \item \textbf{Component criticality threshold}: $\tau_{\text{crit}}$ = 0.05 (5\% relative change in primary metric to deem a component critical)
    \item \textbf{Domain knowledge}: Pre-loaded ablation knowledge base with component-specific guidance
\end{itemize}

For each reproduction task, we first establish a baseline by reproducing the original model, then run \ours to generate and evaluate ablation hypotheses across multiple rounds.

\paragraph{Baseline Configuration and Fairness Guarantees.}
To ensure fair comparison, we configure all baseline methods with consistent settings:

\begin{itemize}
    \item \textbf{LLM Backend}: All methods (Mini-SWE-Agent, RepoMaster, and human experts) use GPT-4o with identical hyperparameters (temperature = 0.7, max tokens = 4096, top-p = 0.95), ensuring consistent generation behavior across systems.
    \item \textbf{Resource Access}: Mini-SWE-Agent and RepoMaster have access to the same repository documentation, README files, and public resources as \ours. However, they do not have access to the pre-built domain knowledge base used by \ours, as this represents domain-specific expertise accumulated through our system design. Notably, RepoMaster's architecture is specifically designed for repository-level search and structure extraction, which compensates for the absence of external domain knowledge and ensures fairness in this comparison. Additionally, we ensure that all task descriptions provided to baseline methods contain sufficiently clear instructions and context, maintaining fair experimental conditions across all systems.
    \item \textbf{Time Budget}: Human experts are given 2 hours per task for closed-book reproduction and ablation design, without access to external documentation beyond the repository itself. This time budget excludes model training execution time, focusing solely on code understanding, modification, and experimental design. This reflects realistic time constraints in code review and verification scenarios.
    \item \textbf{Interaction Steps}: Mini-SWE-Agent and RepoMaster are allowed up to 20 interaction steps per task, sufficient for multi-step debugging and code modification. The one-shot constraint applies to the overall task completion (no restart after failure), not to the number of tool calls within a single execution. In practice, we observe that both baseline methods typically terminate before reaching the maximum iteration limit, often prematurely claiming task completion due to hallucination issues (i.e., the agent incorrectly believes it has successfully completed the task when execution errors or incomplete reproductions remain). This behavior highlights a key challenge in agentic code generation systems and demonstrates the importance of robust verification mechanisms.
\end{itemize}

\paragraph{Justification of One-shot Setting.}
The one-shot per task regime is designed to evaluate each method's ability to autonomously complete reproduction and ablation tasks without human intervention or retry opportunities. This is appropriate for our evaluation because: (1) it reflects real-world automated verification scenarios where systems must handle diverse failure modes autonomously; (2) it tests the robustness of each approach's planning and error recovery mechanisms; (3) both Mini-SWE-Agent and RepoMaster are designed as agentic systems with self-debugging capabilities, making them well-suited for this setting. While this constraint is strict, it fairly assesses each system's end-to-end autonomy and reliability.

\paragraph{Execution Protocol.}
For each reproduction task, we first establish a baseline by reproducing the original model, then run \ours to generate and evaluate ablation hypotheses across multiple rounds. All experiments are conducted on identical hardware (NVIDIA A100 GPUs with 40GB memory) to ensure computational consistency.

\subsection{Candidate Component Space}
\label{appendix:candidate_space}

We provide a detailed breakdown of the candidate component space for each repository:

\input{appendix/tables/candidate_space}

\paragraph{Component Discovery Process.}
Components are identified through a \textbf{hybrid semi-automatic approach}:

\begin{enumerate}
    \item \textbf{Automatic extraction (70-80\%)}: The paper analysis agent parses papers and repository documentation to extract component mentions, architectural descriptions, and module names, producing an initial candidate list.
    
    \item \textbf{Knowledge base validation (20-30\%)}: Extracted components are validated against the pre-built domain knowledge base, which provides guidance on common architectural patterns (e.g., attention mechanisms, GNN layers, normalization) but does \textit{not} pre-specify which components exist in each specific repository.
    
    \item \textbf{Manual filtering (minimal)}: We manually removed 2-3 components per repository that were clearly non-ablatable (\eg data loading utilities, logging functions) to focus on architecturally meaningful ablations.
\end{enumerate}

\paragraph{Search Space Complexity.}
While the number of components (9-15 per repository) may seem modest, the combinatorial nature of mutations results in a substantial number of possible variations. After applying initial filtering to reduce the search space, the remaining components are still subject to multiple mutation types (toggle, scale with various factors, or replacement with alternatives). Each component can mutate in several ways, and combinations of these components can be ablated together. The total number of \textbf{variables} (510, 223, and 193 for CPA, GEARS, and BioLORD respectively) further increases the potential configurations. The role of the bandit algorithm is to efficiently navigate this reduced space to identify the most critical components without exhaustive enumeration, which would be computationally prohibitive for repositories with dozens of configurable components.

\subsection{Model Component Ablation}
\label{appendix:component_impact}

We conduct comprehensive ablation studies to understand the impact of different architectural components. Table~\ref{tab:ablation_study} summarizes the results.

\input{appendix/tables/ablation_results}

\subsubsection{CPA Ablation Results}

For CPA, we explored 7 components over 5 rounds, covering reconstruction loss, adversarial
discriminator, unified/composed latent embedding, encoder network, and perturbation/covariate
embedding dictionaries, as well as dose/time nonlinear scalers. The bandit strategy allocated
most trials to the reconstruction loss and adversarial discriminator arms, which achieved the
highest mean rewards in Table~\ref{tab:ablation_study}. In particular, changing the reconstruction
loss (e.g., to a Negative Binomial variant) and toggling the adversarial branch both produced
systematic but moderate changes in MSE and Pearson correlation, indicating that these two
components are the most leverage points in CPA. In contrast, modifying the encoder network or
the embedding dictionaries led to smaller and less consistent effects, suggesting that the
overall architecture is relatively robust to these design choices.

\subsubsection{GEARS Ablation Results}

For GEARS, we explored 10 components over 5 rounds, including the perturbation and gene GNN
encoders, the combinatorial perturbation aggregator, autofocus loss, GO-derived perturbation
similarity graph, gene coexpression graph, and several decoder and embedding variants.
The combinatorial perturbation aggregator and the perturbation GNN encoder emerged as the most
informative arms with the highest mean rewards, confirming the central role of message passing
and aggregation in GEARS. Ablating the perturbation GNN encoder (bypassing message passing) caused
the strongest degradation in performance, while changes to auxiliary components such as the GO
similarity graph or coexpression graph resulted in smaller, often negligible effects. Overall,
these results highlight that GEARS is most sensitive to how perturbation signals are propagated
and combined, whereas many of the auxiliary graph constructions provide limited additional benefit.

\subsubsection{Biolord Ablation Results}

For BioLORD, we explored 7 latent-space and decoder-related components over multiple rounds.
The most informative hypotheses targeted the parametric output distributions (Gaussian / ZINB / Poisson)
and the unknown-attribute latent embedding $z_u$, which consistently achieved the highest rewards and
slight improvements over the baseline in both MSE and Pearson correlation. In contrast, ablating the
decomposed latent space or completeness loss tended to degrade performance, indicating that explicitly
disentangling known and unknown attributes is important for capturing infection-state variability.
Overall, the ablation study suggests that BioLORD's gains primarily stem from its structured latent
representation and carefully chosen output distributions, while simpler aggregators provide limited benefit.

\subsection{Component Importance Statistical Analysis}
\label{appendix:component_statistics}

This table provides the statistical analysis of the component importance in different tasks, including the \textbf{mean reward}, \textbf{standard deviation}, \textbf{variance}, and \textbf{95\% confidence intervals} (CI) for each component. The analysis is conducted on three main tasks: Biolord, CPA, and GEARS. 
For each task, multiple components are evaluated, and for each component, we report the number of evaluations (\( n \)), the mean reward, the standard deviation (Std Dev), the variance, and the 95\% confidence interval for the component's performance. 

\input{appendix/tables/component_statistics}

The analysis allows us to quantify the relative importance of each component and assess the stability of the ablation process. 
Components with higher variance suggest a greater impact on task performance, highlighting areas that may benefit from further exploration. 
This statistical breakdown not only enhances the understanding of component behavior across tasks but also strengthens the reliability of our conclusions.

%% file: appendix/tables/system_hyperparameters.tex
\begin{table}[ht]
\centering
\caption{Common Hyperparameters and System Settings (All Tasks)}
\label{tab:common_hyperparams}
\small
\begin{tabular}{ll}
\toprule
\textbf{Category} & \textbf{Setting} \\
\midrule
\textbf{Optimization} & \\
\quad Optimizer & Adam (PyTorch default settings) \\
\quad Learning rate & Model-specific defaults from repositories \\
\quad Optimizer parameters & PyTorch default (betas, eps, weight\_decay) \\
\quad Early stopping & Not configured (fixed epoch training) \\
\quad Model checkpointing & Saved after training completion \\
\midrule
\textbf{Data Loading} & \\
\quad DataLoader workers & 0 (single-threaded) \\
\quad Pin memory & Default PyTorch settings \\
\quad Data format & Single-cell gene expression (AnnData/h5ad) \\
\midrule
\textbf{Data Preprocessing} & \\
\quad Normalization & Standard normalization and scaling \\
\quad Split storage & Pre-computed splits in \texttt{data\_split.json} \\
\quad Split strategy & Fixed train/validation/test splits \\
\midrule
\textbf{Evaluation} & \\
\quad Metrics & MSE, MAE, Pearson correlation \\
\quad Inference method & Batch inference \\
\quad Evaluation mode & Model in evaluation mode (no dropout) \\
\midrule
\textbf{Numerical Precision} & \\
\quad Floating point precision & Float32 (PyTorch default) \\
\quad Mixed precision training & Not used \\
\midrule
\textbf{Reproducibility} & \\
\quad Data splits & Pre-computed and stored \\
\quad Model initialization & Repository defaults \\
\quad Environment & Isolated worktree per ablation candidate \\
\quad Version control & All modifications tracked in worktrees \\
\quad Deterministic operations & Model-specific (repository defaults) \\
\midrule
\textbf{Computational} & \\
\quad Device & GPU (CUDA) \\
\quad GPU configuration & CUDA\_VISIBLE\_DEVICES=0 (single GPU) \\
\quad Training mode & Single GPU training \\
\quad Storage & Structured directory organization \\
\bottomrule
\end{tabular}
\end{table}

%% file: appendix/tables/task_parameter.tex
\begin{table*}[ht]
\centering
\caption{Task-Specific Training Parameters}
\label{tab:task_parameters}
\small
\begin{tabular}{lccc}
\toprule
\textbf{Parameter} & \textbf{BioLORD} & \textbf{CPA} & \textbf{GEARS} \\
\midrule
\textbf{Training Configuration} & & & \\
\quad Maximum epochs & 400 & 20 & 20 \\
\quad Batch size & 128 & 1024 & 32 \\
\midrule
\textbf{Data Configuration} & & & \\
\quad Number of cells & 19,053 & 111,122 & 89,357 \\
\quad Number of genes & 8,203 & 5,044 & 5,045 \\
\quad Test set size & 2,859 & 16,669 & 28,754 \\
\quad Train/Val/Test split & 13,337 / 2,857 / 2,859 & 77,785 / 16,668 / 16,669 & $\sim$60,603 / N/A / 28,754 \\
\quad Split key & ``split'' & data\_split.json & data\_split.json \\
\quad Categorical attributes & status, status\_control & gemgroup, tissue\_type, cell\_type & N/A \\
\midrule
\textbf{Baseline Performance} & & & \\
\quad Baseline MSE & 0.2051 & 0.0303 & 0.0044 \\
\quad Baseline Pearson correlation & 0.7303 & 0.9129 & 0.9875 \\
\bottomrule
\end{tabular}%
\end{table*}

%% file: appendix/tables/dataset_statistics.tex
\begin{table}[h]
\centering
\caption{Dataset statistics and sources for the three reproduction tasks (evaluation splits).}
\label{tab:dataset_statistics}
\begin{tabular}{lcccc}
\toprule
\textbf{Task} & \textbf{Type} & \textbf{\#Samples (eval)} & \textbf{\#Genes} & \textbf{Dataset (source)} \\
\midrule
CPA      & single-cell perturbation & $16{,}669$ cells   & $5{,}044$ & Norman2019~\citep{norman2019} \\
GEARS    & single-cell perturbation & $28{,}754$ cells   & $5{,}045$ & Norman2019~\citep{norman2019} \\
BioLORD  & single-cell perturbation & $2{,}859$ cells    & $8{,}203$ & single-cell infection~\citep{piran2024biolord} \\
\bottomrule
\end{tabular}
\end{table}

%% file: appendix/tables/candidate_space.tex
\begin{table}[h]
\centering
\caption{Detailed Candidate Component Space Statistics}
\label{tab:candidate_space}
\begin{tabular}{lccccc}
\toprule
\textbf{Repository} & \textbf{Total} & \multicolumn{3}{c}{\textbf{Mutation Types}} & \textbf{Total} \\
\cmidrule(lr){3-5}
& \textbf{Components} & \textbf{Toggle} & \textbf{Scale} & \textbf{Replace} &  \textbf{Variables}\\
\midrule
CPA     & 12 & 8  & 3 & 1 & 510 \\
GEARS   & 15 & 10 & 4 & 1 & 223\\
BioLORD & 9  & 6  & 2 & 1 & 193 \\
\midrule
\textbf{Total} & \textbf{36} & \textbf{24} & \textbf{9} & \textbf{3} & \textbf{309} \\
\bottomrule
\end{tabular}
\end{table}

%% file: appendix/tables/ablation_results.tex
\begin{table*}[h]

    \centering

    \caption{\textbf{Comprehensive ablation study results showing component-level analysis.} Trials indicates the number of times each component was tested. Reward denotes the online bandit reward used during exploration, and Mean $|\Delta|$ measures the average absolute normalized change in the target performance metric across ablation trials. Metrics show the performance with \textbf{maximum absolute change} (worst-case impact) for each component. Changes are relative to baseline (CPA: MSE=0.0303, Pearson=0.9129; GEARS: MSE=0.0044, Pearson=0.9875; BioLORD: MSE=0.2051, Pearson=0.7303).}

    \label{tab:ablation_study}

    \resizebox{\textwidth}{!}{%

    \begin{tabular}{llcccc}

        \toprule

        \textbf{Task} & \textbf{Component (Arm)} & \textbf{Trials} & \textbf{Reward} & \textbf{MSE (\%$\Delta$, max $|\Delta|$)} & \textbf{Pearson ($\Delta$, max $|\Delta|$)} \\

        \midrule

        \multirow{7}{*}{\textbf{CPA}} & Unified/composed latent embedding & 6 & 4.92 & 0.0395 (\textbf{+30.35\%}) & 0.8854 (\textbf{$-$0.0275}) \\

         & Reconstruction loss & 4 & 5.75 & 0.0335 (\textbf{+10.35\%}) & 0.9034 ($-$0.0094) \\

         & Encoder network & 2 & 2.00 & 0.0327 (+7.89\%) & 0.9056 ($-$0.0073) \\

         & Adversarial discriminator (classifier) & 5 & 5.72 & 0.0295 ($-$2.86\%) & 0.9164 (+0.0035) \\

         & Perturbation embedding dictionary & 3 & 2.67 & 0.0312 (+2.73\%) & 0.9104 ($-$0.0024) \\

         & Covariate embedding dictionary & 3 & 2.67 & 0.0311 (+2.38\%) & 0.9107 ($-$0.0021) \\

         & Dose/time nonlinear scalers & 2 & 1.50 & 0.0304 (+0.30\%) & 0.9126 ($-$0.0003) \\

        \midrule

        \multirow{6}{*}{\textbf{GEARS}} & Combinatorial perturbation aggregator & 4 & 7.58 & 0.0042 (\textbf{$-$4.91\%}) & 0.9878 (+0.0003) \\

         & Learnable gene embeddings & 4 & 4.12 & 0.0043 ($-$1.74\%) & 0.9876 (+0.0001) \\

         & Gene GNN encoder (GNN\_$\theta_g$) & 7 & 6.57 & 0.0050 \textbf{(+14.15\%)} & 0.9864 ($-$0.0011) \\

         & Gene coexpression graph construction & 5 & 3.67 & 0.0043 ($-$1.60\%) & 0.9878 (+0.0003) \\

         & Perturbation GNN encoder (GNN\_$\theta_p$) & 7 & 6.70 & 0.0083 (\textbf{+89.70\%}) & 0.9755 (\textbf{$-$0.0120}) \\

         & GO-derived perturbation similarity graph & 4 & 3.75 & 0.0045 (+2.28\%) & 0.9867 ($-$0.0008) \\

        \midrule

        \multirow{4}{*}{\textbf{BioLORD}} & Unknown-attribute latent embedding ($z_u$) & 7 & 5.25 & 0.3345 (\textbf{+63.05\%}) & 0.6403 (\textbf{$-$0.0900}) \\

         & Known-attribute embeddings & 8 & 3.06 & 0.2161 (\textbf{+5.36\%}) & 0.7183 ($-$0.0119) \\

         & Minimality loss ($\mathcal{L}_\mathrm{min}$) & 3 & 4.50 & 0.2060 (+0.44\%) & 0.7288 ($-$0.0015) \\

         & Classification module (biolord-classify) & 4 & 6.50 & 0.2050 ($-$0.05\%) & 0.7312 (+0.0010) \\

         & Latent aggregator (concatenation) & 3 & 6.10 & 0.2060 (+0.44\%) & 0.7288 ($-$0.0015) \\

        \bottomrule

    \end{tabular}%

    }

\end{table*}

%% file: appendix/tables/component_statistics.tex
\begin{table*}[h]
\centering
\caption{Component Importance Statistical Analysis: Mean Reward with 95\% Confidence Intervals and Variance}
\label{tab:component_statistics}
\small
\begin{tabular}{l l c c c c c}
\toprule
\textbf{Task} & \textbf{Component(Arm)} & \textbf{Trials} & \textbf{Mean} & \textbf{Std Dev} & \textbf{Variance} & \textbf{95\% CI} \\
\midrule
\multirow{5}{*}{\textbf{Biolord}} 
& Unknown-attribute latent embedding ($z_u$) & 7 & 5.250 & 0.945 & 0.893 & [4.375, 6.125] \\
& Known-attribute embeddings & 8 & 3.060 & 0.551 & 0.303 & [2.598, 3.522] \\
& Minimality loss ($\mathcal{L}_{\mathrm{min}}$) & 3 & 4.500 & 0.810 & 0.656 & [2.489, 6.511] \\
& Classification module (biolord-classify) & 4 & 6.500 & 1.170 & 1.369 & [4.640, 8.360] \\
& Latent aggregator (concatenation) & 3 & 6.100 & 1.098 & 1.206 & [3.374, 8.826] \\
\midrule
\multirow{7}{*}{\textbf{CPA}} 
& Unified/composed latent embedding & 6 & 4.920 & 0.886 & 0.784 & [3.991, 5.849] \\
& Reconstruction loss & 4 & 5.750 & 1.035 & 1.071 & [4.104, 7.396] \\
& Encoder network & 2 & 2.000 & 0.360 & 0.130 & [1.976, 2.653] \\
& Adversarial discriminator (classifier) & 5 & 5.720 & 1.030 & 1.060 & [4.440, 7.000] \\
& Perturbation embedding dictionary & 3 & 2.670 & 0.481 & 0.231 & [1.477, 3.863] \\
& Covariate embedding dictionary & 3 & 2.670 & 0.481 & 0.231 & [2.150, 3.675] \\
& Dose/time nonlinear scalers & 2 & 1.500 & 0.270 & 0.073 & [1.114, 2.129] \\
\midrule
\multirow{6}{*}{\textbf{GEARS}} 
& Combinatorial perturbation aggregator & 4 & 7.580 & 1.364 & 1.862 & [5.411, 9.749] \\
& Learnable gene embeddings & 4 & 4.120 & 0.742 & 0.550 & [2.941, 5.299] \\
& Gene GNN encoder (GNN\_$\theta_g$) & 7 & 6.570 & 1.183 & 1.399 & [5.475, 7.665] \\
& Gene coexpression graph construction & 5 & 3.670 & 0.661 & 0.436 & [2.849, 4.491] \\
& Perturbation GNN encoder (GNN\_$\theta_p$) & 7 & 6.700 & 1.206 & 1.454 & [5.583, 7.817] \\
& GO-derived perturbation similarity graph & 4 & 3.750 & 0.675 & 0.456 & [2.677, 4.823] \\
\bottomrule
\end{tabular}%
\end{table*}

%% file: appendix/03_case_studies.tex
\clearpage
\section{Failure Mode Analysis}
\label{sec:failure_analysis}

While \ours achieves high task success rates, we provide a detailed analysis of failure cases observed across CPA, GEARS, and BioLORD repositories to help practitioners understand the system's limitations and inform future improvements. Based on actual execution logs and error patterns, we categorize failures into three main types:

\paragraph{Category 1: Component-to-Code Mapping Failures (Primary Failure Source, $\sim$50\% of failures)}
The majority of failures in our experiments stem from the fragility of mapping paper-level component names to concrete code locations. This manifests in several ways:

\textit{Mismatched component naming}: In BioLORD, multiple ablation attempts targeting "Decomposed latent space" failed with the message "No tracked files were modified after 10 attempts... component name does not match the codebase." The paper-level abstraction refers to latent space decomposition patterns (e.g., \texttt{\_get\_latent\_unknown\_attributes}, \texttt{latent\_unknown\_attributes} fields), but the LLM-based code search could not reliably identify a syntactically editable anchor point despite 10 modification attempts.

\textit{Implicit component dependencies}: Even when components are successfully located, tightly coupled architectural elements pose challenges. For example, ablating graph message-passing layers in GEARS sometimes requires simultaneous modifications to related pooling or aggregation modules, which the system cannot reliably infer from the knowledge base alone.

\textit{Dynamically constructed components}: Components whose structure is determined at runtime (e.g., dynamically built neural network layers, conditional computation paths) are difficult to ablate statically, as their concrete instantiation is not visible in the source code.

This category represents the dominant failure mode: even when paper analysis extracts semantically valid hypotheses, the autonomous mapping from conceptual components to executable code modifications remains brittle. When component names deviate from implementation details, the system conservatively avoids making changes, resulting in failed ablation attempts.

\paragraph{Category 2: Environment and Artifact Issues ($\sim$30\% of failures)}
Contrary to typical dependency conflicts, most environment-related failures in our experiments centered on artifact-level problems rather than basic Python/CUDA setup:

\textit{Corrupted or incompatible model artifacts}: In GEARS, multiple runs failed with "PytorchStreamReader failed reading zip archive: failed finding central directory," indicating corrupted or incompletely downloaded pretrained checkpoint files. These \texttt{torch.load} failures occur after successful environment setup, suggesting that model weight packaging and retrieval is a distinct failure mode.

\textit{Legacy dependency conflicts}: Specific issues include:
\begin{itemize}
    \item \texttt{ImportError: cannot import name 'parse\_use\_gpu\_arg' from 'scvi.model.\_utils'} (API changes in dependencies)
    \item \texttt{ModuleNotFoundError: No module named 'adjustText'/'ray'} (missing optional dependencies)
    \item \texttt{AttributeError: `np.float\_` was removed in NumPy 2.0} (version incompatibilities)
    \item \texttt{TypeError: CPA.setup\_anndata() missing required argument 'control\_group'} (API signature changes)
\end{itemize}

\textit{Data pipeline issues}: Runtime errors such as \texttt{ValueError: num\_samples should be positive but got 0} and \texttt{IndexError: index out of bounds for dimension} indicate empty or malformed data loaders that only surface during training execution.

In controlled environments with template configurations, basic Python/CUDA dependencies are generally resolved successfully. The remaining failures concentrate on pretrained weights validation, legacy API compatibility, and data pipeline robustness—suggesting that future systems need checksum verification for artifacts, automatic fallback mechanisms for corrupted downloads, and more comprehensive API compatibility checks.

\paragraph{Category 3: Code-Level Execution Failures ($\sim$20\% of failures)}
A smaller but significant portion of failures occur when generated ablation code passes syntactic validation but fails during execution:

\textit{Dimension mismatch errors}: In GEARS, ablating certain graph components led to shape mismatches between \texttt{G\_sim} and \texttt{pert\_global\_emb} tensors, indicating that the ablation disrupted downstream tensor operations in ways the system could not anticipate.

\textit{Empty tensor/graph handling}: Missing safety checks for edge cases, such as empty graphs or zero-sized tensors, caused \texttt{IndexError} and \texttt{KeyError} exceptions during training. For example, edge weight creation logic failed when graph structures became degenerate after ablation.

\textit{API compatibility}: Deprecated function calls (e.g., seaborn's \texttt{regplot} parameter changes) and missing exception handling for optional features led to crashes in evaluation scripts.

These failures highlight that semantically meaningful ablations can still introduce subtle bugs that are difficult to detect without dynamic analysis or comprehensive test coverage. The system's current static reasoning cannot predict all runtime consequences of structural changes.

\begin{figure}[h]
    \centering
    \includegraphics[width=0.75\textwidth]{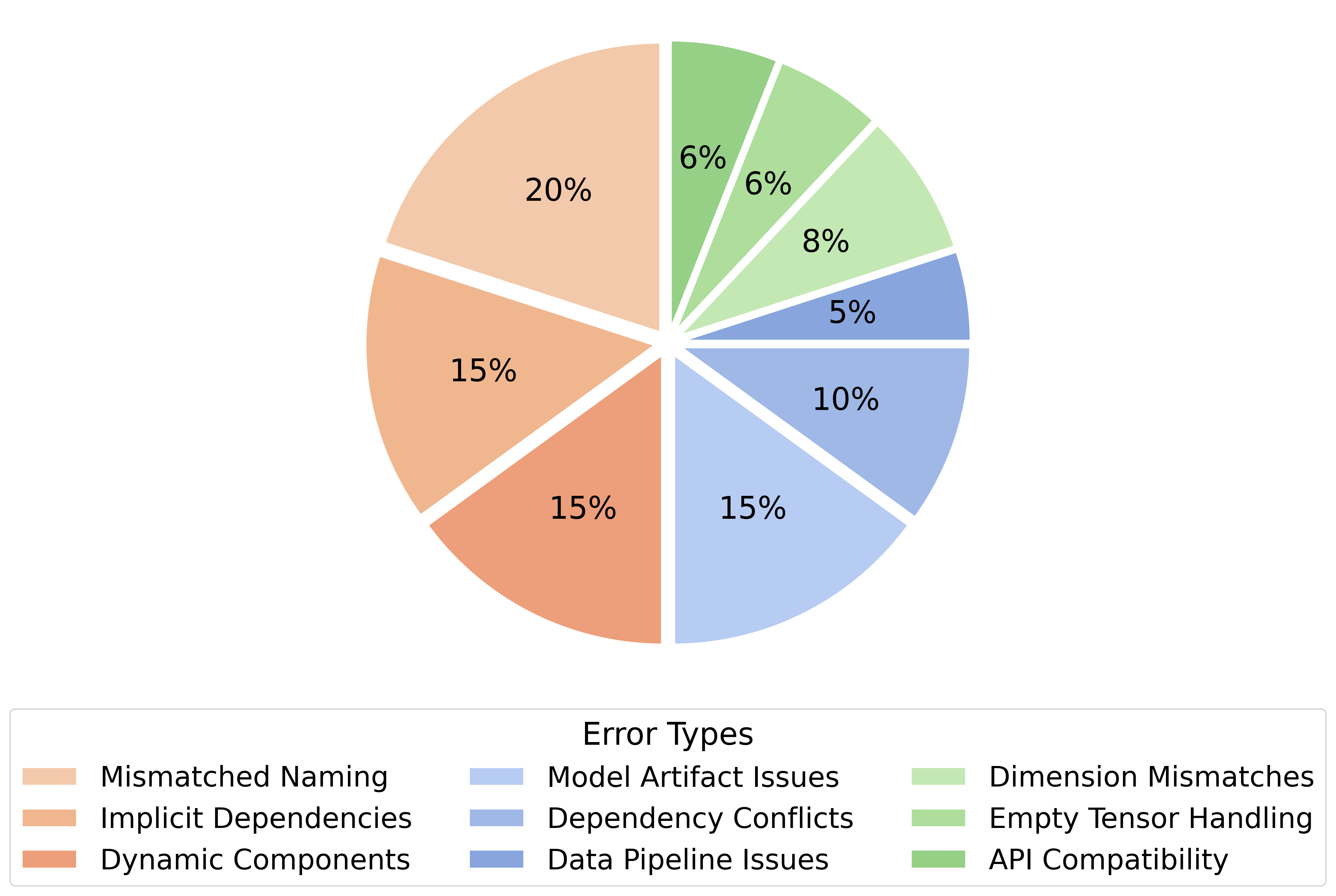}
    \caption{Detailed Distribution of Error Categories}
    \label{fig:error_categories_pie_chart}
\end{figure}

%% file: appendix/04_prompts_and_templates.tex
\clearpage
\section{Prompts}
\label{appendix:prompts}


\begin{figure}[h]
\centering
\begin{minipage}{\linewidth}
\begin{tcolorbox}[
  colback=black!4,
  colframe=black!25,
  fontupper=\normalsize,
  fonttitle=\normalsize\bfseries,
  boxrule=0pt,
  title={Prompt for Hypothesis Agent}
]
\textbf{System Prompt:}

You are a scientific assistant specialized in generating ablation hypotheses from scientific papers
and mapping their described methods to concrete implementations in a target code repository, 
to support reproduction and ablation study planning.

\textbf{Guidelines:}
\begin{itemize}[leftmargin=*,nosep]
\item Understand the paper's structure and key sections (methods, experiments, ablations, implementation details)
\item Identify method components, architectures, hyperparameters, and experimental setups
\item Map paper-described components and procedures to specific code artifacts in the repository (Python modules/classes/functions, configuration files, CLI entrypoints, training/evaluation scripts)
\item Highlight gaps or ambiguities between the paper and the codebase (e.g., missing ablation options)
\item Extract structured information useful for designing ablation studies and configuring experiments
\item Prefer structured, machine-readable outputs when a schema is provided
\item When mapping concepts to code, always reference concrete file paths, symbols, or config keys where possible
\item When uncertain, explain assumptions instead of hallucinating details
\end{itemize}

\vspace{0.6em}
\textbf{Input/Output Specification:}
\begin{itemize}[leftmargin=*,nosep]
\item \textbf{Input}: Paper metadata and codebase context
\item \textbf{Output}: Structured descriptions of method components, experimental setups, and paper-to-code mappings used by downstream Generation Agent for ablation graph execution
\end{itemize}

\end{tcolorbox}
\end{minipage}
\caption{System prompt and interface specification for Hypothesis Agent.}
\label{fig:prompt_paper_analysis}
\end{figure}

\begin{figure}[h]
\centering
\begin{minipage}{\linewidth}
\begin{tcolorbox}[
  colback=black!4,
  colframe=black!25,
  fontupper=\normalsize,
  fonttitle=\normalsize\bfseries,
  boxrule=0pt,
  title={Prompt for Generation Agent}
]
\textbf{System Prompt:}

You are a scientific assistant specialized in generating ablation candidates 
(CandidateSpec) given a candidate space, strategy state, and memory hints.

\textbf{Guidelines:}
\begin{itemize}[leftmargin=*,nosep]
\item Propose meaningful mutations (toggles/scales/replacements/param\_grids) within the candidate space
\item Balance exploration (new components) and exploitation (refining promising ones)
\item Use memory hints and prior results to avoid redundant or obviously bad candidates
\item Keep candidates simple and interpretable
\item Encode mutations in a structured, machine-readable way
\item Favor changes that are feasible to implement and evaluate in the existing codebase
\end{itemize}

\vspace{0.6em}
\textbf{Input/Output Specification:}
\begin{itemize}[leftmargin=*,nosep]
\item \textbf{Input}: Structured candidate space (available components and mutation types), bandit strategy state (arm statistics), and memory hints from previous runs
\item \textbf{Output}: 1--\emph{k} CandidateSpec objects describing concrete mutations (target, action, value, description) to be passed to Code Agent and Ranking Agent for implementation and evaluation
\end{itemize}

\end{tcolorbox}
\end{minipage}
\caption{System prompt and interface specification for Generation Agent.}
\label{fig:prompt_generation}
\end{figure}

\begin{figure}[t]
\centering
\begin{minipage}{\linewidth}
\begin{tcolorbox}[
  colback=black!4,
  colframe=black!25,
  fontupper=\normalsize,
  fonttitle=\normalsize\bfseries,
  boxrule=0pt,
  title={Prompt for Ranking Agent}
]
\textbf{System Prompt:}

You are a scientific analyst specializing in ablation studies and experimental design.
Your task is to compare inference results across multiple ablation study candidates and provide 
detailed comparative analysis.

\textbf{Guidelines:}
\begin{itemize}[leftmargin=*,nosep]
\item Focus on quantitative comparison of inference metrics (mean\_value, std\_value, etc.)
\item Compare each candidate's results against the baseline and other candidates
\item Identify which mutations cause the largest performance changes
\item Look for patterns, trends, and anomalies in the results
\item Assess the relative impact of each mutation on model performance
\item Provide clear, specific findings based on numerical differences
\item Consider both successful and unsuccessful candidates
\item Rank candidates by importance based on actual performance impact
\end{itemize}

Focus on providing actionable insights about which components are most critical based on the 
quantitative differences in inference results.

\vspace{0.6em}
\textbf{Input/Output Specification:}
\begin{itemize}[leftmargin=*,nosep]
\item \textbf{Input}: Candidate-level metrics (including baseline) for one ablation round or full run
\item \textbf{Output}: Ranked list of candidates with explanations of relative impact, consumed by Reflection Agent and Analysis Agent
\end{itemize}
\end{tcolorbox}
\end{minipage}
\caption{System prompt and interface specification for Ranking Agent.}
\label{fig:prompt_ranking}
\end{figure}

\begin{figure}[t]
\centering
\begin{minipage}{\linewidth}
\begin{tcolorbox}[
  colback=black!4,
  colframe=black!25,
  fontupper=\normalsize,
  fonttitle=\normalsize\bfseries,
  boxrule=0pt,
  title={Prompt for Reflection Agent}
]
\textbf{System Prompt:}

You are a scientific analyst specializing in ablation studies and experimental design.
Your task is to analyze ablation study results and provide strategic feedback to guide future candidate generation.

\textbf{Guidelines:}
\begin{itemize}[leftmargin=*,nosep]
\item Analyze patterns in candidate results to identify critical components
\item Assess the effectiveness of the current ablation strategy
\item Identify mutations/components with the most significant impact
\item Provide actionable recommendations for improving the ablation study
\item Consider both successful and unsuccessful candidates
\item Suggest strategy adjustments for better insights
\item Be specific and data-driven in your analysis
\end{itemize}

Your goal is to optimize the ablation study by identifying the most informative experiments to run next.

\vspace{0.6em}
\textbf{Input/Output Specification:}
\begin{itemize}[leftmargin=*,nosep]
\item \textbf{Input}: Ranked candidate results (across multiple rounds) and current bandit state
\item \textbf{Output}: Strategy recommendations and updated priorities that guide Generation Agent and Planner Agent in subsequent rounds
\end{itemize}
\end{tcolorbox}
\end{minipage}
\caption{System prompt and interface specification for Reflection Agent.}
\label{fig:prompt_reflection}
\end{figure}

\begin{figure}[t]
\centering
\begin{minipage}{\linewidth}
\begin{tcolorbox}[
  colback=black!4,
  colframe=black!25,
  fontupper=\normalsize,
  fonttitle=\normalsize\bfseries,
  boxrule=0pt,
  title={Prompt for Analysis Agent}
]
\textbf{System Prompt:}

You are a scientific analyst responsible for aggregating and interpreting ablation results.

\textbf{Guidelines:}
\begin{itemize}[leftmargin=*,nosep]
\item Analyze candidate-level metrics and statuses across the full ablation run
\item Identify global trends, best configurations, and failure patterns
\item Produce concise, structured summaries and recommendations
\item Focus on patterns that would influence future experimental design
\item Distinguish clearly between strong evidence and weak/inconclusive signals
\item When possible, connect quantitative findings to concrete configuration changes
\end{itemize}

\vspace{0.6em}
\textbf{Input/Output Specification:}
\begin{itemize}[leftmargin=*,nosep]
\item \textbf{Input}: All ablation runs for a task (metrics, statuses, rankings, reflections)
\item \textbf{Output}: Global summary describing key components, informative failures, and recommended
      configurations, typically written into the final report
\end{itemize}

\end{tcolorbox}
\end{minipage}
\caption{System prompt and interface specification for Analysis Agent.}
\label{fig:prompt_analysis}
\end{figure}

\begin{figure}[t]
\centering
\begin{minipage}{\linewidth}
\begin{tcolorbox}[
  colback=black!4,
  colframe=black!25,
  fontupper=\normalsize,
  fonttitle=\normalsize\bfseries,
  boxrule=0pt,
  title={Prompt for Planner Agent}
]
\textbf{Intent classifier prompt:}\\
You are an intent classifier for Coding tasks. Return ONLY a JSON object with keys: \\
wants\_train (bool), wants\_infer (bool), wants\_plot (bool), dataset\_key (string or null), model\_key (string or null), dataset\_path (string or null), model\_dir (string or null), task\_name (string or null), step\_keywords (object mapping step name - \> list of keywords). \\
Allowed step names: Load data, Setup data, Load pretrained model, Train model, Run inference, Save/plot outputs. \\
If the task includes explicit paths (e.g., .h5ad or Pretrained\_Models/\dots), return them as dataset\_path/model\_dir. \\
Available dataset keys: \{datasets\_csv\} \\
Available model keys: \{models\_csv\} \\
Task: \{task\}

\vspace{0.6em}

\textbf{Template selector prompt:}\\
You are selecting which templates to include for an agent instruction. \\
Return ONLY a JSON array of filenames in the order they should appear. \\
You must choose from the available filenames. If unsure, include more rather than fewer.

Task:\\
\{task\}

Available templates:\\
\{template\_lines\}
\end{tcolorbox}
\end{minipage}
\caption{System and user prompt for Planner Agent}
\label{fig:prompt_planner}
\end{figure}

\begin{figure}[t]
\centering
\begin{minipage}{\linewidth}

\begin{tcolorbox}[
  colback=black!4,
  colframe=black!25,
  fontupper=\normalsize,
  fonttitle=\normalsize\bfseries,
  boxrule=0pt,
  title={Prompt for Coding Agent}
]

\textbf{System template:}\\
You are a helpful assistant that can interact with a computer. \\

Your response must contain exactly ONE bash code block with ONE direct command (no chaining with \texttt{\&\&}, \texttt{||}, or \texttt{;}). \\
You may use heredocs (e.g., \texttt{cat << EOF > file.py \dots EOF}) to write files when needed. \\
Include a THOUGHT section before your command where you explain your reasoning process. \\
Format your response as shown in \texttt{<format\_example>}. \\

\texttt{<format\_example>}\\
Your reasoning and analysis here. Explain why you want to perform the action. \\

\texttt{bash}\\
\texttt{your\_command\_here} \\

\texttt{</format\_example>} \\

Failure to follow these rules will cause your response to be rejected.

\vspace{0.6em}

\textbf{Instance template:}\\
Please solve this issue: \{task\} \\

You can execute bash commands and edit files to implement the necessary changes.

\end{tcolorbox}

\end{minipage}
\caption{System and user prompt for Code Agent}
\label{fig:prompt_code}
\end{figure}